\documentclass[10pt,twocolumn,letterpaper]{article}

\usepackage[pagenumbers]{cvpr} 

\usepackage{graphicx}
\usepackage{amsmath}
\usepackage{amssymb}
\usepackage{booktabs}
\usepackage{caption}
\usepackage{epsfig}
\usepackage{multirow}
\usepackage[table]{xcolor}
\usepackage{tabularx, etoolbox}

\usepackage[pagebackref,breaklinks,colorlinks]{hyperref}
\definecolor{yellowgreen}{HTML}{98CC70}

\usepackage[capitalize]{cleveref}
\crefname{section}{Sec.}{Secs.}
\Crefname{section}{Section}{Sections}
\Crefname{table}{Table}{Tables}
\crefname{table}{Tab.}{Tabs.}

\def\cvprPaperID{4944} 
\def\confName{CVPR}
\def\confYear{2023}

\begin{document}

\title{DP-NeRF: Deblurred Neural Radiance Field with Physical Scene Priors}
\author{Dogyoon Lee$^{1}$ \quad
Minhyeok Lee$^{1}$ \quad
Chajin Shin$^{1}$ \quad
Sangyoun Lee$^{1,2}$ \\
\vspace{-0.1cm}
$^{1}$Yonsei University\\
$^{2}$Korea Institute of Science and Technology (KIST)\\
{\tt\small \{nemotio, hydragon516, chajin, syleee\}@yonsei.ac.kr}
}


\maketitle

\begin{abstract}
\vspace{-0.2cm}
Neural Radiance Field (NeRF) has exhibited outstanding three-dimensional (3D) reconstruction quality via the novel view synthesis from multi-view images and paired calibrated camera parameters. 
However, previous NeRF-based systems have been demonstrated under strictly controlled settings, with little attention paid to less ideal scenarios, including with the presence of noise such as exposure, illumination changes, and blur.
In particular, though blur frequently occurs in real situations, NeRF that can handle blurred images has received little attention.
The few studies that have investigated NeRF for blurred images have not considered geometric and appearance consistency in 3D space, which is one of the most important factors in 3D reconstruction.
This leads to inconsistency and the degradation of the perceptual quality of the constructed scene.
Hence, this paper proposes a DP-NeRF, a novel clean NeRF framework for blurred images, which is constrained with two physical priors.
These priors are derived from the actual blurring process during image acquisition by the camera.
DP-NeRF proposes rigid blurring kernel to impose 3D consistency utilizing the physical priors and adaptive weight proposal to refine the color composition error in consideration of the relationship between depth and blur. 
We present extensive experimental results for synthetic and real scenes with two types of blur: camera motion blur and defocus blur.
The results demonstrate that DP-NeRF successfully improves the perceptual quality of the constructed NeRF ensuring 3D geometric and appearance consistency.
We further demonstrate the effectiveness of our model with comprehensive ablation analysis.
\end{abstract}
\vspace{-0.7cm}
\footnote{Code: \textit{\url{https://github.com/dogyoonlee/DP-NeRF}}}
\footnote{Project: \textit{\url{https://dogyoonlee.github.io/dpnerf/}}}

\section{Introduction}
\label{sec:intro}
\vspace{-0.2cm}
The synthesis of the photo-realistic novel view image of complex three-dimensional (3D) scenes has advanced rapidly due to the emergence of the Neural Radiance Field(NeRF)~\cite{mildenhall2020nerf}.
NeRF has introduced implicit scene representation to the field, which maps an arbitrary continuous 3D coordinate to the volume density and radiance color using volume-rendering technique and implicit neural representation.
NeRF densely reconstructs continuous 3D space to produce photorealistic rendered images with novel view. 

Though NeRF has achieved remarkable success in a variety of fields, most of the NeRF variants have been designed and tested for a carefully controlled environment that requires well-captured images from multiple views with calibrated camera parameters.
However, various forms of noises are usually included in the data captured for the NeRF in real scenarios, complicating geometric and appearance consistency in 3D representation.

Several NeRF variants have attempted to reconstruct 3D scene in the presence of noise, including exposure noise~\cite{mildenhall2022nerfinthedark,huang2022hdrnerf,jun2022hdrplenoxel}, motion~\cite{park2021nerfies,park2021hypernerf,li2021nsff,pumarola2021dnerf,tretschk2021nonrigid,li2022neural3dvideo,lombardi2019neural,zhang2021editable}, illumination changes~\cite{martin2021nerfinthewild,chen2022hallucinated,zhang2021nerfactor}, and  aliasing~\cite{barron2021mipnerf,barron2022mipnerf360}.
\begin{figure}[t!]
      \begin{center}
         \includegraphics[scale=1.0]{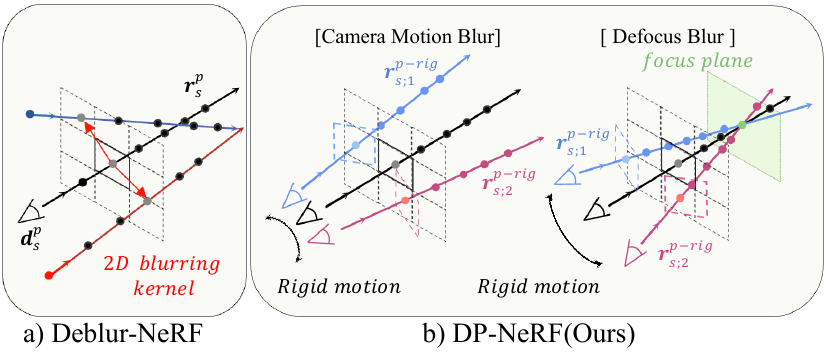} 
      \vspace{-0.7cm}
      \caption{(a) Deblur-NeRF models blurring kernel based on 2D offset on the image pixels. This modeling break the consistency in trained neural radiance field due to the lack of a 3D consistency priors. However, (b) DP-NeRF can render clean neural radiance field guaranteeing the 3D consistency with rigid motion of the camera based on the physical priors of the blur occurrence.}
      \label{fig:representative_fig}
      \end{center}
	  \vspace{-1cm}
\end{figure}
However, although it frequently occurs in real-world settings, blur has not been sufficiently addressed to date, despite the fact that it generates critical artifacts in 3D scene reconstruction.
Deblur-NeRF~\cite{ma2022deblurnerf} introduced blurring kernel estimation for a NeRF by imitating in-camera blurred image acquisition based on a blind deblurring method. 
Their method demonstrated excellent performance and produced clearly rendered images from multi-view images.
However, the blurring kernel in \cite{ma2022deblurnerf} is implemented by optimizing ray deformation and composition weights depending on the 2D pixel location independently, leading to insufficient 3D information.
In reality, the blurring process occurs simultaneously for all pixels in an image due to the physical process of in-camera image acquisition, but \cite{ma2022deblurnerf} overlooks the prior for blurring, leading to a lack of consistency in an image.
Moreover, the designed kernel can be inherently optimized to subobtimal in regions with complex depth or similar appearance due to the independent optimization of the deformation of each ray.
As a result, the estimated kernel has difficulty aggregating 3D information in a way that guarantees geometric and appearance consistency.

In this paper, we propose a deblurred NeRF based on two physical scene priors (hereafter, \textbf{DP-NeRF}) with a novel rigid blurring kernel (RBK) and adaptive weight proposal (AWP). 
The RBK consists of rigid ray transformation (RRT) and coarse composition weights (CCW), which utilize explicit physical scene priors derived from the blurring process to construct a consistent 3D scene representation from blurred images. 
In addition, the AWP proposes fine-grained color composition weights considering the relationship between depth and blur to create more realistic and clean 3D representation.
Furthermore, we propose coarse-to-fine optimization for stable training and to gradually increase the effect of the AWP during training by introducing exponential weight decay between the two losses from the RBK and AWP.
\figurename~\ref{fig:representative_fig} summarizes the DP-NeRF's system using the rigid motion of the camera.

The RBK generates a 3D deformation field and coarse weights for color composition based on the view information for each scene regardless of the pixel for each ray.
This architecture is inspired by the physical scene prior that the blurring process consistently occurs for all pixels for a specific view. 
Specifically, the deformation field is constructed as the 3D rigid motion of the camera for each view and does not depend on the 2D spatial position of each ray.
In contrast to Deblur-NeRF~\cite{ma2022deblurnerf}, our model successfully models 3D space with consistent geometry and appearance due to the use of these conditional physical priors and not fully depending on 2D pixel-wise independent ray optimization. 

Previous studies have claimed that color composition process in a blurring kernel are affected by the depth values of the pixels when compositing blurred colors from both camera motion and defocus blur~\cite{srinivasan2017light,srinivasan2018aperture}.
Hence, RBK can lose detail in regions that have a complex depth or similar textures even though it achieves remarkably realistic 3D scene.
For this reason, the AWP refines the composition weights using feature modulation (FM)~\cite{zhao2022nerfattention} and novel motion feature aggregation module(MAM) based on the depth features of samples for transformed rays, the viewing direction, and the view information. 
Following the ~\cite{ma2022deblurnerf}, we jointly optimize the RBK, AWP, and sharp NeRF with only the reconstruction loss from the blurred input as supervision.
During inference stage, we can clearly render a reconstructed 3D scene using only the trained sharp NeRF model.

The rest of the paper is structured as follows.
In Section~\ref{sec:method}, we describe the RBK and AWP in detail.
In Section~\ref{subsec:exp_NVS} and supplementary material, we provide experimental results for novel view synthesis using synthetic and real scene datasets with two types of blur that are provided from~\cite{ma2022deblurnerf}.
The results show that DP-NeRF achieves significant quantitative and qualitative improvement, preserving 3D consistency with a cleanly rendered novel view.
In addition, we extensively analyze the effectiveness of the proposed model in Section~\ref{subsec:exp_abl}. 
We also demonstrate how the RBK approximately models the blurring process in the supplementary material.
To summarize, this paper offers the following major contributions.

\vspace{-0.2cm}	
\begin{itemize}
	\item \textit{Rigid blurring kernel.} We propose a novel RBK to construct a clean NeRF from blurred images utilizing physical scene priors derived from the blurring process during image acquisition.
	\vspace{-0.2cm}	
	\item \textit{Adaptive weight proposal.} We propose an AWP to refine the composition weights in the RBK considering the relationship between depth and blur to generate more realistic results.
	\vspace{-0.2cm}	
	\item \textit{Coarse-to-fine optimization.} To fully utilize proposed methods in training, we propose coarse-to-fine optimization by applying exponential weight decay between the reconstruction loss from the RBK and AWP.
	\vspace{-0.6cm}	
	\item \textit{Significant improvement in perceptual quality.} DP-NeRF produces enhanced 3D scene representation with greater perceptual quality and clean photo-realistic rendered images.
\end{itemize}
\vspace{-0.5cm}	

\section{Related work}
\label{sec:rework}
\vspace{-0.1cm}
\noindent \textbf{NeRF under various conditions.} 
NeRF has become widespread in computer vision and graphics tasks related to neural rendering, utilizing coordinate-based implicit neural representation (INR). 
Due to the success of the NeRF in neural rendering, several studies have applied NeRF to other areas such as dynamic scenes~\cite{park2021nerfies,li2021nsff,pumarola2021dnerf,tretschk2021nonrigid,park2021hypernerf,li2022neural3dvideo,lombardi2019neural,zhang2021editable}, generative models~\cite{schwarz2020graf,niemeyer2021giraffe}, relighting~\cite{srinivasan2021nerv,bi2020neuralreflectappear,martin2021nerfinthewild,philip2021free}, human avatars~\cite{zhao2022humannerf,su2021anerf,peng2021animatablenerf}, and 3D reconstruction~\cite{wang2021neus,sun2021neuralrecon}.
However, few studies have been conducted under non-ideal conditions~\cite{barron2021mipnerf,barron2022mipnerf360,mildenhall2022nerfinthedark,ma2022deblurnerf,huang2022hdrnerf,jun2022hdrplenoxel}. 
Mip-NeRF~\cite{barron2021mipnerf} addresseed the aliasing issue of ray samples by introducing 3D conical frustum ray casting with integrated positional encoding.
Mip-NeRF 360~\cite{barron2022mipnerf360} then extended Mip-NeRF~\cite{barron2021mipnerf} to unbounded 360-degree scenes using shrunken space parametrization and online distillation to improve its quality and efficiency.
To address the inconsistent appearance and transient objects in the uncarefully collected images, NeRF-W~\cite{martin2021nerfinthewild} introduced appearance and transient latent codes to the NeRF.
HDR-NeRF~\cite{huang2022hdrnerf} and HDR-Plenoxel~\cite{jun2022hdrplenoxel} modeled the high dynamic range (HDR) radiance, imitating the physical process of in-camera image acquisition. 
\cite{huang2022hdrnerf} modeled camera response function with the exposure value for the NeRF and \cite{jun2022hdrplenoxel} modeled white balance function for Plenoxels~\cite{yu2021plenoxels}.
Deblur-NeRF~\cite{ma2022deblurnerf} explored a new area of research constructing a clean NeRF from blurred images, which regularly occur during image acquisition in real-scenario. 

\noindent \textbf{Image Deblurring.} 
Blur can be categorized into four types: camera motion, defocus, moving object and mixed blur. 
Image deblurring aims to recover a sharp image from images degraded by these types of blur.
The recovery process can be expressed as to solve the equation: $B=I*K$, where $B$, $I$, and $K$ denote the blurred image, sharp image, and blurring kernel, respectively.
Deblurring can be divided into two categories, non-blind and blind, whose difference is whether the blurring kernel is known or not.
Recent studies have focused primarily on blind deblurring because the blurring kernel is typically unknown in real-scenarios.

Several traditional image deblurring techniques~\cite{shan2008highmotiondeblur, jia2007transparencydeblur, joshi2009deblurcolorpirors, krishnan2011blind} have been proposed based on maximum a posterior (MAP) estimation~\cite{richardson1972bayesian,lucy1974iterative} based on a prior condition derived from natural images as a form of regularization.
\cite{shan2008highmotiondeblur} uses global and local image priors as two piece-wise continuous functions and local smoothness constraints, while \cite{jia2007transparencydeblur} proposes generalized transparency to efficiently estimate the blur filter by selecting useful pixels based on new transparency map.
\cite{joshi2009deblurcolorpirors, krishnan2011blind} propose a sparse prior derived from local color statistics and a regularization function as the ratio of the l1-norm to the l2-norm for the high frequencies of an image, respectively.

Recently, deep image deblurring has been investigated following the success of deep learning networks in computer vision field. 
In this approach, a general blurring kernel is usually employed and latent images constructed based on blind deblurring and a data-driven prior through network training with paired datasets.
Several studies have been proposed the use of convolutional neural network(CNN)~\cite{sun2015learningcnndeblur, nah2017deepdeblurring,tao2018scale,zhang2019deephier,zamir2021multi} and generative models~\cite{kupyn2018deblurgan, kupyn2019deblurganv2}.
We focus on priors from traditional methods because physical priors are helpful for constructing a blurring kernel in a NeRF system.

\noindent \textbf{NeRF from Blurred Images.} 
Deblur-NeRF~\cite{ma2022deblurnerf} models the blurring kernel with the NeRF imitating the blind deblurring to produce clean and sharp NeRF.
In contrast to recent deblurring methods that operate in the image space, the target blur types are camera motion and defocus blur in a static scene. 
Motion blur is excluded as a separate problem that needs to be overcome because temporal inconsistency is another challenge in 3D reconstruction.
In ~\cite{ma2022deblurnerf}, blurring kernel is optimized based on the kernel points on the 2D image pixels and view-embedded information.
The kernel is designed around transformed rays that penetrate the kernel points on the image plane and camera origin, which are independently optimized during the training.
However, the kernel relies heavily on the training of the deep neural network without cues for geometric and appearance consistency in 3D scene representation, which leads to a lack of consistency in 3D scene.
Our method focuses on this limitation and proposes a novel blurring kernel with two physical priors derived from physical process of blur acquisition and ray casting as a form of regularization for kernel estimation.
\begin{figure*}
		\includegraphics[width=2.1\columnwidth]{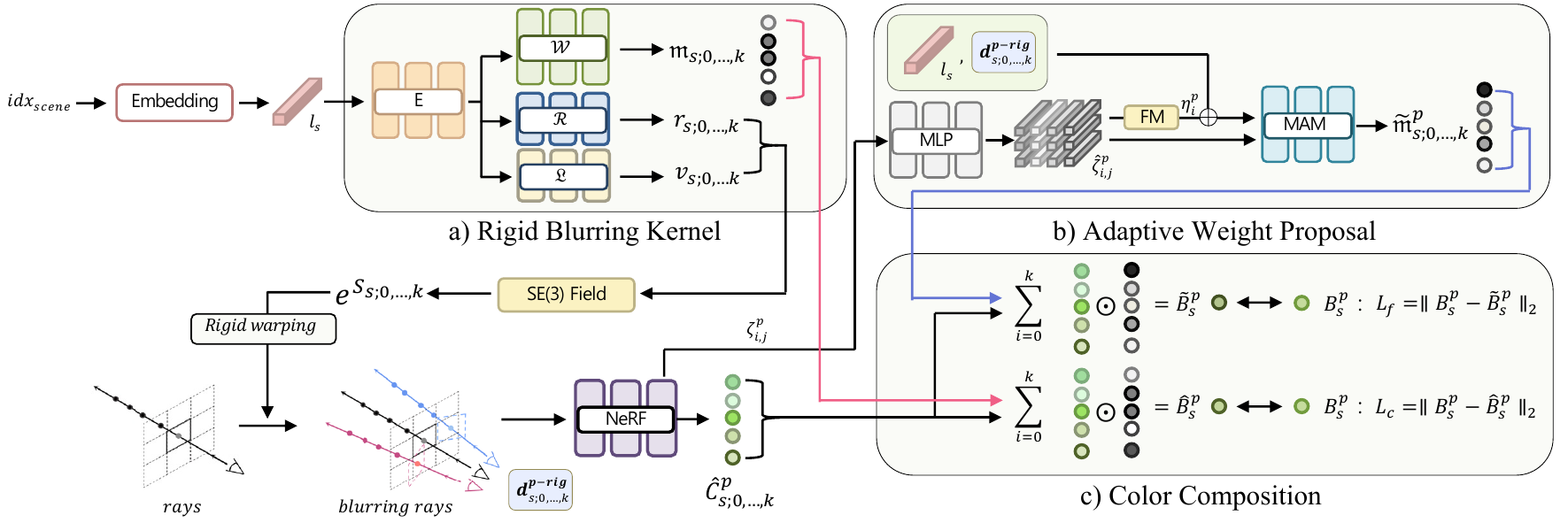}
      \vspace{-0.7cm}
      \caption{Overall pipeline for DP-NeRF. DP-NeRF consists of three stages. (a) The rigid blurring kernel (RBK) constructs the blurring system using the $SE(3)$ Field based on the physical priors. (b) The adaptive weight proposal (AWP) refines the composition weights using the depth feature ($\zeta^{p}_{i,j}$) of the samples on the ray of the target pixel ($p$), the scene ($s$) information, and the rigidly transformed ray directions ($\textbf{r}^{p}_{s;0,\dots,k}$). (c) Finally the coarse and fine blurred colors, $\hat{B}^{p}$ and $\tilde{B}^{p}$, are composited using the weighted sum of the ray transformed colors. $\mathcal{L}_{c}$ and $\mathcal{L}_{f}$ denote the coarse and fine RGB reconstruction loss, respectively.}
      \vspace{-0.55cm}
      \label{fig:DP-NeRF_pipeline}
   \end{figure*}
\vspace{-0.5cm}
\section{Deblurred Neural Radiance Field}
\label{sec:method}
In this section, we describe our process for constructing a clean NeRF given a set of blurred inputs.
Initially, we model the RBK to use the blur consistency in an image as a physical scene priors(Section~\ref{subsec:RBK}).
To consider the relationship between depth and blur, we then model the AWP module(Section~\ref{subsec:AWP}).
Finally, we explain our loss function and coarse-to-fine optimization strategy for the training of DP-NeRF(Section~\ref{subsec:Optim}).
Overall process for DP-NeRF is summarized in~\figurename~\ref{fig:DP-NeRF_pipeline}.

\subsection{Preliminary}
\label{subsec:prelim}
\noindent \textbf{Neural Radiance Field (NeRF).} 
NeRF~\cite{mildenhall2020nerf} constructs a continuous, volumetric representation of a 3D scene based on INR. 
It uses a multi layer perceptron(MLP) to approximate the function 
\vspace{-0.1cm}
\begin{equation} \label{eq:nerf}
F:\left(\gamma_{\textbf{x}}(\textbf{x}),\gamma_{\textbf{d}}(\textbf{d})\right)\rightarrow(\textbf{c},\sigma),
\end{equation}
which maps 3D position $\textbf{x}=(x,y,z)$ and viewing direction $\textbf{d}=(\phi,\theta)$ to a color $c=(r,g,b)$ and volume density $\sigma$.

Specifically, the 3D position and viewing direction are independently projected to a higher dimension by applying the sinusoidal positional encoding function $\gamma:\mathbb{R}^{3}\rightarrow\mathbb{R}^{3+6m}$, which is defined as
\vspace{-0.1cm}
\begin{equation} \label{eq:pos_enc}
\gamma(\textbf{x})=\left(\textbf{x},\dots,sin(2^{f}\pi\textbf{x}),cos(2^{f}\pi\textbf{x}),\dots\right), 
\vspace{-0.1cm}
\end{equation}
where $f=\{0,\dots,m-1\}$ and $m$ is a hyper-parameter that decides the frequency band.
For clarity, we abbreviate the positional encoding and represent the NeRF as
\vspace{-0.1cm}
\begin{equation} \label{eq:nerf_abbr}
F(\textbf{x},\textbf{d}) = (\textbf{c},\sigma).
\vspace{-0.1cm}
\end{equation}
To train the NeRF with input images, the NeRF renders each color $\hat{C}^{p}$ of pixel $p$ using a rendering technique~\cite{max1995opticalraytracing} that is an approximated version of classical volume rendering~\cite{kajiya1984raytracing}.
For given ray origin $\textbf{o}^{p}$ and viewing direction $\textbf{d}^{p}$ along a pixel $p$, the $i_{th}$ sample on the ray $\textbf{r}^{p}$ is defined as $\textbf{r}_{i}^{p}=\textbf{o}^{p}+t_{i}\textbf{d}^{p}$, where $t_{i}$ is drawn from $N$ evenly spaced bins with stratified sampling~\cite{mildenhall2020nerf} in near-to-far bounded partition $[t_{n}, t_{f}]$ as shown in Eq.~\ref{eq:vol_sample}:
\begin{equation} \label{eq:vol_sample}
t_{i}\sim\mathcal{U}\left[t_{n}+\frac{i-1}{N}\left(t_{f}-t_{n}\right),t_{n}+\frac{i}{N}\left(t_{f}-t_{n}\right)\right].
\end{equation}
Pixel color $\hat{C}^{p}$ is computed from the predicted color $\textbf{c}_{i}^{p}$ and density $\sigma_{i}^{p}$ of each sample $\textbf{r}_{i}^{p}$  as shown in Eq.~\ref{eq:vol_ren}:
\vspace{-0.2cm}
\begin{equation} \label{eq:vol_ren}
\hat{C}(\textbf{r})=\sum_{i=1}^N w_{i}\textbf{c}_{i}=\sum_{i=1}^N T_{i}\left(1-exp(-\sigma_{i}\delta_{i})\right)\textbf{c}_{i},
\vspace{-0.2cm}
\end{equation}
where $T_{i}=exp(-\sum_{j=1}^{i-1}\sigma_{j}\delta_{j})$ is the transmittance and $\delta_{i}=t_{i+1}-t_{i}$ is the distance between adjacent samples.
Note that, we abbreviate the notation for pixel $p$ for clarity.

\noindent \textbf{Blind Deblurring in the NeRF.} 
Our goal is to solve the blind deblurring process with respect to sharp pixel color $\hat{C}^{p}$ and blurring kernel $h^{p}$ in a similar manner to~\cite{ma2022deblurnerf} as 
\vspace{-0.2cm}
\begin{equation} \label{eq:blind_deblur}
\hat{B}^{p}=\hat{C}^{p}*h^{p},
\vspace{-0.2cm}
\end{equation}
where $\hat{C}^{p}$ is computed from sample color $\textbf{c}_{i}^{p}$ and density $\sigma_{i}^{p}$, which are predicted by the NeRF.
\cite{ma2022deblurnerf} models kernel $h^{p}$ by introducing approximated $n$ sparse kernel points in $K \times K$ sized window $\mathcal{N}(p)$ at the 2D pixel location, which are optimized by the MLP.
The rendered pixel colors from the kernel points are then composited by the corresponding weights $w^{p}_q$, which are also predicted by the MLP, as Eq.~\ref{eq:deblurnerf_eq}:
\vspace{-0.1cm}
\begin{equation} \label{eq:deblurnerf_eq}
\hat{B}^{p}=\sum_{q\in\mathcal{N}(p)}w^{p}_{q}\hat{C}^{p}_{q}
\vspace{-0.1cm}
\end{equation}
The kernel points and weights for each pixel ray are optimized independently depending on the 2D spatial pixel coordinates and view-information.
Thus, the blurring kernel is dynamic, but they overlook the importance of geometric and appearance consistency in overall 3D space.
\vspace{-0.1cm}

\subsection{Rigid Blurring Kernel (RBK)}
\label{subsec:RBK}
\vspace{-0.1cm}
\noindent \textbf{Physical Scene Priors.}
As we mentioned in Section~\ref{subsec:prelim}, deblurring in the NeRF should consider 3D consistency. 
To address this, we impose two priors as constraints inspired by the physical process of image blurring.

\textbf{\textit{Prior 1: A blurred image is generated during in-camera image acquisition.}}
The first prior shares ray rigid transformation (RRT) from camera motion through all of the pixels of a blurred image because same camera is used.

\textbf{\textit{Prior 2: The blurring process for all of the pixels in a blurred image occurs simultaneously.}}
The second prior shares the coarse composition weights (CCW) across all of the pixels of an image because the color composition of all of the pixels in a blurred image is affected simultaneously.

\noindent \textbf{Ray Rigid Transformation.}
From the first prior, we mimic the blurring process of an image using RRT based on each image's view information.
RRT is formulated as ray transformation derived from the deformation of rigid camera motion, defined as the dense $SE(3)$ field for scene $s$, approximated by the MLPs, which consists of shared encoder MLP $E$  and independent MLPs ($\mathcal{R}$ and $\mathfrak{L}$), as shown in Eq.~\ref{eq:se3field}:
\vspace{-0.2cm}
\begin{equation} \label{eq:se3field}
\mathcal{S}_{s}=\big( \mathcal{R}(E(l_{s})); \mathfrak{L}(E(l_{s})) \big),~~where~~s\in N_{img},
\vspace{-0.2cm}
\end{equation}
where $l_{s}$ denotes the latent code for each scene through the embedding layer~\cite{bojanowski2017latentoptimizing}, and $N_{img}$ denotes a set of image indices.
The scene-wise $SE(3)$ field can encode the rigid motion of the camera for the scene, thus consistently transforming the rays of the scene to imitate the blurring process.
Inspired by Nerfies~\cite{park2021nerfies}, rigid motion is encoded as the screw axis~\cite{lynch2017modernrobotics} $\mathcal{S}_{s}=(r_{s};v_{s})\in\mathbb{R}^{6}$, where $r_{s}\in \mathfrak{so}(3)$ encodes rotation. $\hat{r}_{s}=r_{s}/||r_{s}||$ is the axis of rotation and $\theta=||r_{s}||$ is the angle of rotation.
Rotation matrix  $e^{r_{s}}\in SO(3)$ is taken from Rodrigues' formula~\cite{rodrigues1816attraction}:
\vspace{-0.2cm}
\begin{equation} \label{eq:Rodrigeus_rotation}
e^{r_{s}}\equiv e^{[r_{s}]}=\textbf{I}+\frac{\sin\theta}{\theta}[r_{s}]_{\times}+\frac{1-\cos\theta}{\theta^{2}}[r^{2}_{s}]_{\times},
\vspace{-0.1cm}
\end{equation}
where $[\mathbf{x}]_{\times}$ denotes the cross-product matrix of vector $\mathbf{x}$.
The translation matrix, which is encoded by screw motion $\mathcal{S}_{s}$, is taken as $\textbf{p}_{s}=\textbf{G}_{s}v_{s}$, where
\vspace{-0.1cm}
\begin{equation} \label{eq:Rodrigeus_translation}
\textbf{G}_{s}=\textbf{I}+\frac{1-\cos\theta}{\theta^{2}}[r_{s}]_{\times}+\frac{\theta-\sin\theta}{\theta^{3}}[r_{s}]^{2}_{\times}.
\end{equation}
For given ray $\textbf{r}^{p}_{s}$ on arbitrary pixel $p$ in scene $s$, we can define the RRT as
\vspace{-0.2cm}
\begin{equation} \label{eq:RRT}
\textbf{r}^{p-rig}_{s;q} = e^{\mathcal{S}^{p}_{s;q}}\textbf{r}^{p}_{s}=e^{\mathcal{S}_{s;q}}\textbf{r}^{p}_{s}=e^{r_{s;q}}\textbf{r}^{p}_{s}+\textbf{p}_{s;q},
\vspace{-0.2cm}
\end{equation}
where $q\in\{1,\dots,k\}$, $k$ is a hyper-parameter that controls the number of camera motions contributing to the blur in scene $s$, and $\textbf{r}^{p-rig}_{s;q}$ denotes the rigidly transformed (RT) ray from $\textbf{r}^{p}_{s}$.
Note that, $\mathcal{S}^{p}_{s;q} = \mathcal{S}_{s;q}$ due to our first prior.
To ensure that, for $\mathfrak{se}(3)$, $e^{\mathcal{S}_{s;q}}$ is the identity when $\mathcal{S}_{s;q}=0$, we initialize the weights of the last layer of MLP $\mathcal{R}$ from $\mathcal{U}(-10^{-5},10^{-5})$ following \cite{park2021nerfies}.
The transformed sharp colors $\hat{C}^{p-rig}_{s;q}$ are then rendered from $\textbf{r}^{p-rig}_{s;q}$ using a volume-rendering technique to composite the blurry color $\hat{B}^{p}_{s}$ based on the CCW described in the following section. 

\noindent \textbf{Coarse Composition Weights.}
From the second prior, we model the CCW field for scene $s$ from the MLP $\mathcal{W}$, which shares the encoding MLP $E$ with other MLPs ($\mathcal{R}$ and $\mathfrak{L}$).
\vspace{-0.2cm}
\begin{equation} \label{eq:composition_field}
\mathfrak{m}_{s;0,\dots,k} = \sigma\left(\mathcal{W}(E(l_{s}))\right),~~where~~\sum^{k}_{i=0}\mathfrak{m}_{s;i}=1,
\end{equation}
where $k$ denotes the number of camera motions shared with the RRT. $\mathfrak{m}_{s}$ is the CCW for scene $s$. $\sigma$ represents the sigmoid function.
Note that, the number of $\mathfrak{m}_{s}$ is $k+1$, where $\mathfrak{m}_{s;0}$ is the weight for original ray $\textbf{r}^{p}_{s}$.
Finally, blurry color~$\hat{B}^{p}_{s}$~for pixel $p$ in scene $s$, is composited by the weighted sum of the rendered colors of original ray $\hat{C}^{p}_{s;0}$ and RT rays $\hat{C}^{p-rig}_{s;1,\dots,k}$ using the corresponding per-scene CCW $\mathfrak{m}_{s;0,\dots,k}$ as shown in Eq.~\ref{eq:coarse_color_composition}:
\vspace{-0.2cm}
\begin{equation} \label{eq:coarse_color_composition}
\hat{B}^{p}_{s} = \mathfrak{m}_{s;0}\hat{C}^{p}_{s;0} + \sum_{i=1}^{k}\mathfrak{m}_{s;i}\hat{C}^{p-rig}_{s;i}
\end{equation}
The RBK pipeline is summarized in \figurename~\ref{fig:DP-NeRF_pipeline}~(a) and (c).

\subsection{Adaptive Weight Proposal (AWP)}
\label{subsec:AWP}
\noindent \textbf{Relationship between Depth and Blur.}
Though the proposed RBK models the blurring kernel successfully with realistic rendering results (\figurename~\ref{fig:qualitative_results_synthetic}), there is still room for improvement in the relationship between depth and blur.
Several past studies have described the relationship between depth and each type of blur~\cite{srinivasan2017light,srinivasan2018aperture}.
Specifically, camera motion blur is affected by depth when the camera motion is out-of-plane~\cite{srinivasan2017light}, while defocus blur is usually affected by depth~\cite{srinivasan2018aperture}.
Therefore, we employ the AWP module to alleviate the color composition errors by flexibly refining the CCW $\mathfrak{m}_{s;0,\dots,k}$ along the depth features of the samples on the original and RT rays for pixel $p$.
Further description for motivation of AWP is attached in supplementary material.

\begin{figure}[t!]
      \begin{center}
         \includegraphics[scale=1.0]{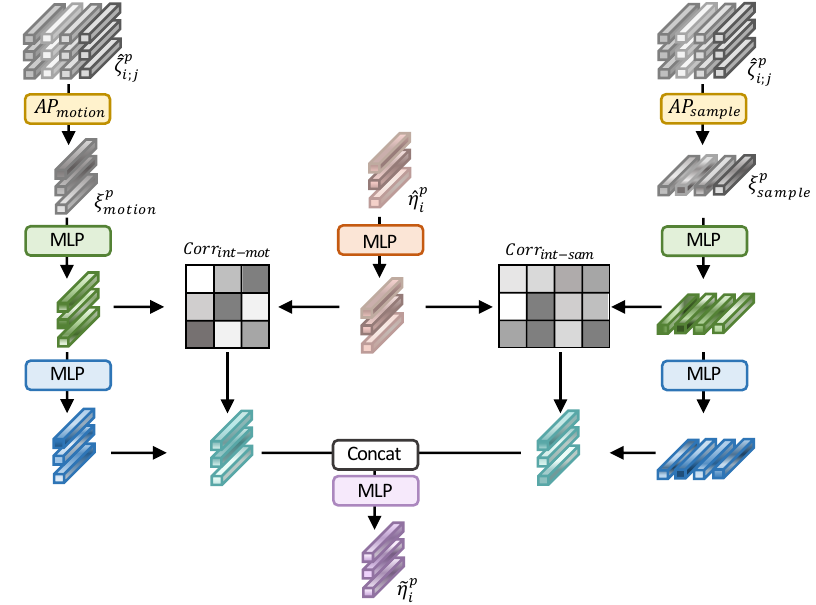} 
  	  \vspace{-0.8cm}
      \caption{Architecture of the motion aggregation module (MAM).}
      \label{fig:mam_architecture}
      \end{center}
	  \vspace{-0.8cm}
\end{figure}
\noindent \textbf{AWP Network.}
We define the AWP as a function that infers the per-pixel adaptive composition weights $\tilde{\mathfrak{m}}^{p}_{s;0,\dots,k}$ utilizing the depth features of each sample on the rays, corresponding to the latent code of each scene, and the direction of each ray.
The function $\mathcal{AWP}$ is defined as Eq.~\ref{eq:awp}:
\begin{equation} \label{eq:awp}
\tilde{\mathfrak{m}}^{p}_{s;0,\dots,k} = \mathcal{AWP}(\zeta^{p}_{i=0,\dots,k;j=1,\dots,N},\textbf{d}^{p-rig}_{i=0,\dots,k},l_{s}),
\end{equation}
where $i$ denotes the original and RT rays, and $j$ denotes the sample on each ray.
In addition, $\zeta$ denotes the corresponding depth feature and $\textbf{d}^{p-rig}_{0,\dots,k}$ denotes the directions of original and RT rays.
We then composite the adaptive blurred color $\tilde{B}^{p}_{s}$ through $\tilde{\mathfrak{m}}^{p}_{s;0,\dots,k}$ as shown in Eq.~\ref{eq:awp_composition}:
\vspace{-0.2cm}
\begin{equation} \label{eq:awp_composition}
\tilde{B}^{p}_{s} = \tilde{\mathfrak{m}}^{p}_{s;0}\hat{C}^{p}_{s;0} + \sum_{i=1}^{k}\tilde{\mathfrak{m}}^{p}_{s;i}\hat{C}^{p-rig}_{s;i}
\vspace{-0.2cm}
\end{equation}
We design $\mathcal{AWP}$ to be an approximation using a deep learning network (\figurename~\ref{fig:DP-NeRF_pipeline}~(b) and (c)).

\noindent \textbf{Network Architecture.}
Inspired by CurveNet~\cite{xiang2021walk}, the AWP generates the fine weights $\tilde{\mathfrak{m}}^{p}_{s;i}$ utilizing the inter-sample and inter-motion correlation of the rays transformed by the RBK, which are denoted as $Corr_{int-sam}$ and $Corr_{int-mot}$ in \figurename~\ref{fig:mam_architecture}, respectively.
The depth features for each sample on the rays~($\zeta^{p}_{i;j}$) are extracted from second-to-last layer of the NeRF, which contains implicit occupancy information.
First, we construct motion-wise modulated features, $\eta^{p}_{i=0,\dots,k}$, through feature modulation~(FM)~\cite{zhao2022nerfattention} from $N$ samples on each ray as shown in \figurename~\ref{fig:DP-NeRF_pipeline}~(b) and Eq.~\ref{eq:feature_modulation}:
\vspace{-0.2cm}
\begin{equation} \label{eq:feature_modulation}
\eta^{p}_{i} = \sum^{N}_{l=1}\big(\exp(-\sum^{l-1}_{m=1}\delta_{m}\hat{\zeta}^{p}_{i;m}\circ(1-\exp(-\delta_{l}\hat{\zeta}^{p}_{i;l})\circ\hat{\zeta}^{p}_{l})\big),
\end{equation}
where $\hat{\zeta}^{p}_{i;j}$ denotes the embedded features of $\zeta^{p}_{i;j}$ from the simple MLP and $\delta$ is the same as in Eq.~\ref{eq:vol_ren}. Second, we add the viewing direction and scene-information by using a simple MLP with $\eta^{p}_{i}$, $\textbf{d}^{p-rig}_{i=0,\dots,k}$, and $l_{s},$ as shown in Eq.~\ref{eq:added_modulated}:
\vspace{-0.2cm}
\begin{equation} \label{eq:added_modulated}
\hat{\eta}^{p}_{i} = MLP\left(\eta^{p}_{i}, \gamma_{\textbf{d}}(\textbf{d}^{p-rig}_{i}), l_{s}\right),~~where~~i\in \{0,\dots,k\},
\end{equation}
where $\hat{\eta}^{p}_{i}$ denotes the modulated features with the viewing information. 
We then forward ($\zeta^{p}_{i;j}$ and $\hat{\eta}^{p}_{i}$) to the MAM to aggregate the implicit depth-derived information based on attentive feature extraction as in Eq.~\ref{eq:mam}:
\vspace{-0.2cm}
\begin{equation} \label{eq:mam}
\tilde{\eta}^{p}_{i} = MAM(\hat{\zeta}^{p}_{i;j}, \hat{\eta}^{p}_{i}),
\vspace{-0.2cm}
\end{equation}
where $\tilde{\eta}^{p}_{i}$ denotes the aggregated features.

For the MAM, we first generate motion-wise and sample-wise representative features, $\xi^{p}_{motion}$ and $\xi^{p}_{sample}$, by forwarding the embedding MLP and employing attentive pooling~\cite{hu2020attentivepooling} along each axis from the embedded features $\hat{\zeta}^{p}_{i;j}$. Note that, we omit the embedding MLP in \figurename~\ref{fig:mam_architecture} for clarity.
Second, we compute the inter-motion and inter-sample correlation using CurveNet-like architecture via matrix multiplication.
Third, the correlated features update each embedded features using matrix multiplication.
Finally, aggregated features $\tilde{\eta}^{p}_{i}$ are extracted by concatenating the correlated embedded features and forwarding the simple MLP.
The overall process for the MAM is described as \figurename~\ref{fig:mam_architecture} and Eq.~\ref{eq:mam_detail}:
\vspace{-0.1cm}
\begin{equation} \label{eq:mam_detail}
MAM(\hat{\zeta}^{p}_{i;j}, \hat{\eta}^{p}_{i})=MLP\big(cat(\textbf{corr}(\hat{\eta}^{p}_{i}, \xi^{p}_{motion}, \xi^{p}_{sample}))\big),
\end{equation}
where $cat$ denotes the concatenating operation and $\textbf{corr}$ represents computing operations of inter-motion and inter-sample correlation.
To this end, the adaptive composition weights $\tilde{\mathfrak{m}}^{p}_{s;0,\dots,k}$ are predicted using global average pooling (GAP) along the motion axis and the linear layer as shown in Eq.~\ref{eq:adaptive_weight}.
We omit the GAP and the final Linear layer for clarity in \figurename~\ref{fig:DP-NeRF_pipeline}~(b).
\vspace{-0.5cm}
\begin{equation} \label{eq:adaptive_weight}
\tilde{\mathfrak{m}}^{p}_{s;i}=\sigma\big(Linear(GAP(\tilde{\eta}^{p}_{i=0,\dots,k}))\big),~~where~~\sum^{k}_{i=0}\tilde{\mathfrak{m}}^{p}_{s;i}=1
\end{equation}
Details of the operation and notations for the feature dimensions are presented in the supplementary material.

\subsection{Training \& Optimization}
\label{subsec:Optim}
\noindent \textbf{Training Loss.}
DP-NeRF takes only the RGB reconstruction loss for the blurred color of a pixel with a corresponding ray because our goal is to approximate the blurred color of the pixel using the RBK and AWP.
In contrast to Deblur-NeRF~\cite{ma2022deblurnerf}, we employ two blurred colors $\hat{B}^{p}_{s}$ and $\tilde{B}^{p}_{s}$ to optimize the DP-NeRF.
Our RGB reconstruction loss $\mathcal{L}_{recon}$ consists of two reconstruction losses from these predicted colors and the ground truth color as shown in Eq.~\ref{eq:training_loss}:
\vspace{-0.1cm}
\begin{equation} \label{eq:training_loss}
\mathcal{L}_{recon} = ||B^{p}_{s}-\hat{B}^{p}_{s}||+||B^{p}_{s}-\tilde{B}^{p}_{s}||,
\vspace{-0.1cm}
\end{equation}
where $B^{p}_{s}$ and $\mathcal{L}$ denote the ground truth blurred RGB for pixel $p$ and the loss function, respectively.

\noindent \textbf{Coarse-to-Fine Optimization.}
However, it is difficult to simultaneously optimize the loss function from scratch due to the complex geometry and texture of 3D scenes.
Hence, we propose a coarse-to-fine optimization strategy for the two losses by introducing coarse-to-fine weight $\lambda$, which exponentially decays from $\lambda_{s}$ to $\lambda_{e}$ during training as shown in Eq.~\ref{eq:lambda_decay}:
\vspace{-0.3cm}
\begin{align} \label{eq:lambda_decay}
\begin{split}
\alpha & = -log(\lambda_{s}/\lambda_{e})/(e_{f}-e_{c}),\\
\lambda & = \lambda_{s}(\exp(\alpha(e_{c}-e_{i}))),
\end{split}
\end{align}
where $e_{c}$, $e_{i}$, and $e_{f}$ denote the current, initial, and final iteration of the training process, respectively.
Therefore, our final loss function $\mathcal{L}_{final}$ is defined as shown in Eq.~\ref{eq:final_loss}:
\begin{equation} \label{eq:final_loss}
\mathcal{L}_{final} = \lambda||B^{p}_{s}-\hat{B}^{p}_{s}||+(1-\lambda)||B^{p}_{s}-\tilde{B}^{p}_{s}||,
\end{equation}
where the first and second terms without $\lambda$ and $(1-\lambda)$ are denoted as $\mathcal{L}_{c}$ and $\mathcal{L}_{f}$ in the caption of \figurename~\ref{fig:DP-NeRF_pipeline}, referring to coarse and fine RGB reconstruction loss, respectively.

\section{Experiment}
\label{sec:exp}
\noindent \textbf{Implementation Details.}
DP-NeRF is implemented using official code published for a previous work~\cite{ma2022deblurnerf}.
Note that, our blur operation should be applied to scene irradiance instead of image intensity, as pointed out by \cite{chen2012theoretical}, following \cite{ma2022deblurnerf}.
Hence, a tone mapping function is applied to the predicted radiance color from the NeRF in the same manner as the gamma function in \cite{ma2022deblurnerf}.
For fair comparison with \cite{ma2022deblurnerf}, we set the default configuration to be same as  in that study.
The number of camera motions $k$ is set to $4$ as the default because the number of kernel points in \cite{ma2022deblurnerf} is set to $5$.
We use a batch size of 1024 rays, with 64 coarse samples, and 64 fine samples on the rays.
We use the Adam~\cite{kingma2014adam} optimizer with default parameters. 
For the scheduling learning rate, we exponentially weight decay from $5\times10^{-4}$ to $8\times10^{-5}$.
In addition, we set $\lambda_{s}$ and $\lambda_{e}$ at $0.9$ and $0.1$ for the coarse-to-fine optimization, respectively.
We also use $200k$ iterations to train each scene. Further details are provided in the supplementary material.

\noindent \textbf{Datasets.}
We train DP-NeRF using the synthetic and real scene datasets provided by~\cite{ma2022deblurnerf}.
Both dataset consist of two blur types: camera motion and defocus blur.
There are five scenes in the synthetic dataset and ten in the real dataset for each blur type.
The camera poses for all of the images are calibrated using COLMAP~\cite{schonberger2016pixelwise,schonberger2016structure}.
As mentioned in~\cite{ma2022deblurnerf}, the real scenes were manually captured with a Canon EOS RP under manual exposure mode.
\begin{table*}
   \Large
   \begin{center}
      \resizebox{2.1\columnwidth}{!}{
		\centering
		\setlength{\tabcolsep}{1pt}
         \begin{tabular}{c||ccc|ccc|ccc|ccc|ccc|ccc}
	        \specialrule{0.6pt}{1pt}{1pt}
			\multirow{2}{*}{Camera Motion}& \multicolumn{3}{c}{Factory} & \multicolumn{3}{c}{Cozyroom} & \multicolumn{3}{c}{Pool} & \multicolumn{3}{c}{Tanabata} & \multicolumn{3}{c}{Trolley} & \multicolumn{3}{c}{Average} \\
			 & PSNR($\uparrow$) & SSIM($\uparrow$) & LPIPS($\downarrow$) & PSNR($\uparrow$) & SSIM($\uparrow$) & LPIPS($\downarrow$) & PSNR($\uparrow$) & SSIM($\uparrow$) & LPIPS($\downarrow$) & PSNR($\uparrow$) & SSIM($\uparrow$) & LPIPS($\downarrow$) & PSNR($\uparrow$) & SSIM($\uparrow$) & LPIPS($\downarrow$) & PSNR($\uparrow$) & SSIM($\uparrow$) & LPIPS($\downarrow$) \\
			\midrule
			Naive NeRF~\cite{mildenhall2020nerf} & 19.32 & .4563 & .5304 & 25.66 & .7941 & .2288 & 30.45 & .8354 & .1932 & 22.22 & .6807 & .3653 & 21.25 & .6370 & .3633 & 23.78 & .6807 & .3362 \\
			MPR~\cite{zamir2021multi}~+~NeRF & 21.70 & .6153 & .3094 & 27.88 & .8502 & .1153 & 30.64 & .8385 & .1641 & 22.71 & .7199 & .2509 & 22.64 & .7141 & .2344 & 25.11 & .7476 & .2148 \\
			PVD~\cite{son2021pvd}~+~NeRF & 20.33 & .5386 & .3667 & 27.74 & .8296 & .1451 & 27.56 & .7626 & .2148 & 23.44 & .7293 & .2542 & 23.81 & .7351 & .2567 & 24.58 & .7190 & .2475 \\
			Deblur-NeRF~\cite{ma2022deblurnerf} & \cellcolor{yellow!25}25.60 & \cellcolor{yellow!25}.7750 & \cellcolor{yellow!25}.2687 & \cellcolor{yellow!25}32.08 & \cellcolor{yellow!25}.9261 & \cellcolor{yellow!25}.0477 & \cellcolor{yellow!25}31.61 & \cellcolor{yellow!25}.8682 & \cellcolor{yellow!25}.1246 & \cellcolor{yellow!25}27.11 & \cellcolor{yellow!25}.8640 & \cellcolor{yellow!25}.1228 & \cellcolor{yellow!25}27.45 & \cellcolor{yellow!25}.8632 & \cellcolor{yellow!25}.1363 & \cellcolor{yellow!25}28.77 & \cellcolor{yellow!25}.8593 & \cellcolor{yellow!25}.1400 \\
			DP-NeRF &  \cellcolor{orange!25}25.91 & \cellcolor{orange!25}.7787 & \cellcolor{orange!25}.2494 & \cellcolor{orange!25}32.65 & \cellcolor{orange!25}.9317 & \cellcolor{orange!25}.0355 & \cellcolor{orange!25}31.96 & \cellcolor{orange!25}.8768 & \cellcolor{orange!25}.0908 & \cellcolor{orange!25}27.61 & \cellcolor{orange!25}.8748 & \cellcolor{orange!25}.1033 & \cellcolor{orange!25}28.03 & \cellcolor{orange!25}.8752 & \cellcolor{orange!25}.1129 & \cellcolor{orange!25}29.23 & \cellcolor{orange!25}.8674 & \cellcolor{orange!25}.1184 \\
			\midrule\midrule
			\multirow{2}{*}{Defocus}& \multicolumn{3}{c}{Factory} & \multicolumn{3}{c}{Cozyroom} & \multicolumn{3}{c}{Pool} & \multicolumn{3}{c}{Tanabata} & \multicolumn{3}{c}{Trolley} & \multicolumn{3}{c}{Average} \\
			 & PSNR($\uparrow$) & SSIM($\uparrow$) & LPIPS($\downarrow$) & PSNR($\uparrow$) & SSIM($\uparrow$) & LPIPS($\downarrow$) & PSNR($\uparrow$) & SSIM($\uparrow$) & LPIPS($\downarrow$) & PSNR($\uparrow$) & SSIM($\uparrow$) & LPIPS($\downarrow$) & PSNR($\uparrow$) & SSIM($\uparrow$) & LPIPS($\downarrow$) & PSNR($\uparrow$) & SSIM($\uparrow$) & LPIPS($\downarrow$) \\
			\midrule
			Naive NeRF~\cite{mildenhall2020nerf} & 25.36 & .7847 & .2351 & 30.03 & .8926 & .0885 & 27.77 & .7266 & .3340 & 23.80 & .7811 & .2142 & 22.67 & .7103 & .2799 & 25.93 & .7791 & .2303 \\
			KPAC~\cite{son2021kpac}~+~NeRF & 26.40 & .8194 & .1624 & 28.15 & .8592 & .0815 & 26.69 & .6589 & .2631 & 24.81 & .8147 & .1639 & 23.42 & .7495 & .2155 & 25.89 & .7803 & .1773 \\
			Deblur-NeRF~\cite{ma2022deblurnerf} & \cellcolor{yellow!25}28.03 & \cellcolor{yellow!25}.8628 & \cellcolor{yellow!25}.1127 & \cellcolor{yellow!25}31.85 & \cellcolor{yellow!25}.9175 & \cellcolor{yellow!25}.0481 & \cellcolor{yellow!25}30.52 & \cellcolor{yellow!25}.8246 & \cellcolor{yellow!25}.1901 & \cellcolor{yellow!25}26.26 & \cellcolor{yellow!25}.8517 & \cellcolor{yellow!25}.0995 & \cellcolor{yellow!25}25.18 & \cellcolor{yellow!25}.8067 & \cellcolor{yellow!25}.1436 & \cellcolor{yellow!25}28.37 & \cellcolor{yellow!25}.8527 & \cellcolor{yellow!25}.1188 \\
			DP-NeRF & \cellcolor{orange!25}29.26 & \cellcolor{orange!25}.8793 & \cellcolor{orange!25}.1035 & \cellcolor{orange!25}32.11 & \cellcolor{orange!25}.9215 & \cellcolor{orange!25}.0386 & \cellcolor{orange!25}31.44 & \cellcolor{orange!25}.8529 & \cellcolor{orange!25}.1563 & \cellcolor{orange!25}27.05 & \cellcolor{orange!25}.8635 & \cellcolor{orange!25}.0779 & \cellcolor{orange!25}26.79 & \cellcolor{orange!25}.8395 & \cellcolor{orange!25}.1170 & \cellcolor{orange!25}29.33 & \cellcolor{orange!25}.8713 & \cellcolor{orange!25}.0987 \\
	        \specialrule{0.6pt}{1pt}{1pt}
			\bottomrule
      \end{tabular}
      }%
   \vspace{-0.1cm}
   \caption{Quantitative results for the synthetic scene. Each color shading indicates the \colorbox{orange!25}{best} and \colorbox{yellow!25}{second-best} result, respectively.}
   \label{tab:quantitative_results_synthetic}
   \end{center}
   \vspace{-0.4cm}
\end{table*}
\begin{figure*}
         \includegraphics[scale=1.0]{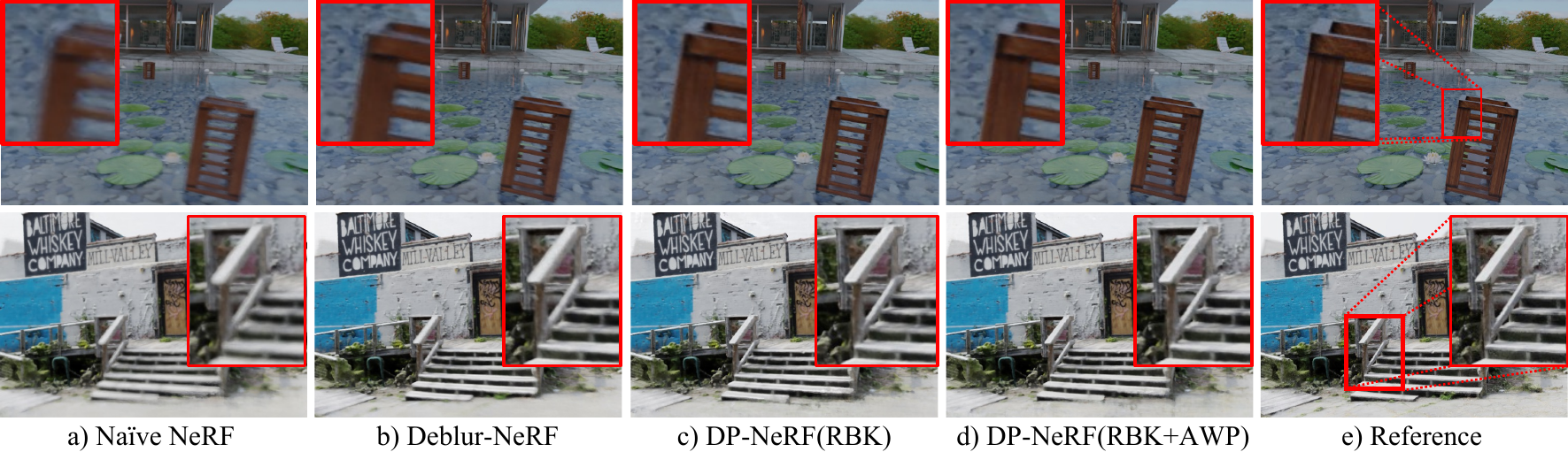} 
	   \vspace{-0.5cm}
      \caption{Rendered novel view synthesis results of DP-NeRF for synthetic scenes. Top and bottom row denote results of camera motion and defocus blur scene, respectively. \figurename~(a)-(e) denote Naive NeRF, Deblur-NeRF, DP-NeRF (RBK), DP-NeRF (RBK+AWP), and ground truth images, respectively. Each colored box in corner of images are enlarged parts of the colored box region in reference images.}
      \vspace{-0.3cm}
      \label{fig:qualitative_results_synthetic}
   \end{figure*}

\begin{table}
   \Huge
   \begin{center}
      \resizebox{\columnwidth}{!}{
		\setlength{\tabcolsep}{1pt}
         \begin{tabular}{c|ccc||c|ccc}
			\multicolumn{4}{c}{Camera Motion} & \multicolumn{4}{c}{Defocus}\\
	        \specialrule{0.6pt}{1pt}{1pt}
			 & ~PSNR($\uparrow$)~ & SSIM($\uparrow$)~ & LPIPS($\downarrow$)~ & & ~PSNR($\uparrow$)~ & SSIM($\uparrow$)~ & LPIPS($\downarrow$) \\
			\midrule
			Naive NeRF~\cite{mildenhall2020nerf}~ & 22.69 & .6347 & .3687 & ~Naive NeRF~\cite{mildenhall2020nerf}~ & 22.40 & .6661 & .2310 \\
			MPR~\cite{zamir2021multi}~+~NeRF~ & 23.38 & .6655 & .3140 & ~KPAC~\cite{son2021kpac}~+~NeRF~ & 23.04 & .6917 & .1847 \\
			PVD~\cite{son2021pvd}~+~NeRF~ & 23.10 & .6389 & .3425 & - & - & - & - \\
			Deblur-NeRF~\cite{ma2022deblurnerf}~ & \cellcolor{yellow!25}25.63 & \cellcolor{yellow!25}.7675 & \cellcolor{yellow!25}.1820 & ~Deblur-NeRF~\cite{ma2022deblurnerf}~ & \cellcolor{yellow!25}23.46 & \cellcolor{yellow!25}.7199 & \cellcolor{yellow!25}.1207 \\
			DP-NeRF~ & \cellcolor{orange!25}25.91 & \cellcolor{orange!25}.7751 & \cellcolor{orange!25}.1602 & ~DP-NeRF~ & \cellcolor{orange!25}23.67 & \cellcolor{orange!25}.7299 & \cellcolor{orange!25}.1082 \\
	        \specialrule{0.6pt}{1pt}{1pt}
			\bottomrule
      \end{tabular}
      }
   \vspace{-0.3cm}
   \caption{Average results for the real scene dataset. Each color shading indicates the \colorbox{orange!25}{best} and \colorbox{yellow!25}{second-best} result, respectively.}
   \label{tab:average_quantitative_results_real}
   \end{center}
   \vspace{-0.5cm}
\end{table}
\subsection{Novel View Synthesis}
\label{subsec:exp_NVS}
\noindent \textbf{Evaluations.}
In this section, we summarize the results of our model for the synthetic and real scenes.
The quantitative quantitative and qualitative results for the synthetic dataset are presented in \tablename~\ref{tab:quantitative_results_synthetic} and \figurename~\ref{fig:qualitative_results_synthetic}.
Three commonly used evaluation metrics are adopted in the present study to compare the synthesized and ground truth images: the peak signal-to-noise ratio (PSNR), the structural similarity index measure (SSIM), and learned perceptual image patch similarity (LPIPS)~\cite{zhang2018lpips}, which assess relative sharpness, structural similarity, and perceptual quality, respectively.
Due to the length, we only present the average results here for the real scene dataset.
A more version of the experimental results is available in the supplementary material.
\vspace{0.1cm}

\noindent \textbf{Comparisons.}
Tables~\ref{tab:quantitative_results_synthetic} and \ref{tab:average_quantitative_results_real} show that our model produces excellent results for all metrics compared to the other models, including \cite{ma2022deblurnerf}.
In particular, LPIPS is significantly improved by DP-NeRF, indicating a 3D scene with a higher rendered image quality in terms of perceptual quality.
Note that, the results for single image deblurring methods, MPR+NeRF, PVD+NeRF, and KPAC+NeRF, in Tables~\ref{tab:quantitative_results_synthetic} and \ref{tab:average_quantitative_results_real} are taken from \cite{ma2022deblurnerf}.
They are trained with a Naive NeRF using images deblurred using MPR~\cite{zamir2021multi}, PVD~\cite{son2021pvd}, and KPAC~\cite{son2021kpac} methods, respectively.

The qualitative results presented in \figurename~\ref{fig:qualitative_results_synthetic} demonstrate the effectiveness of DP-NeRF.
The RBK more successively models a clean 3D scene representation with geometric and appearance consistency compared to previous approaches. 
It more cleanly reproduces the structure of the object in the scene compared to Naive NeRF and Deblur-NeRF.
\vspace{-0.2cm}

\subsection{Ablation Study}
\label{subsec:exp_abl}
\noindent \textbf{Effectiveness of the RBK and AWP.}
The DP-NeRF successfully constructs clean NeRF with geometric and appearance consistency using RBK and AWP as shown in \figurename~\ref{fig:qualitative_results_synthetic}.
The RBK by itself still has difficulty in inferring the correct texture and geometry in some regions in which the texture is confused with the background, the structure is thin, or the depth is complex, as indicated by the red box in \figurename~\ref{fig:qualitative_results_synthetic}~(c).
However, the rendered results in \figurename~\ref{fig:qualitative_results_synthetic}~(d) demonstrate that the AWP effectively refines the geometric and appearance consistency in this region.
In particular, the upper and lower rows exhibit detailed enhancement of the geometric structure and texture of the edge of the box and stairs, respectively.

\begin{figure}[t!]
      \begin{center}
         \includegraphics[scale=1.0]{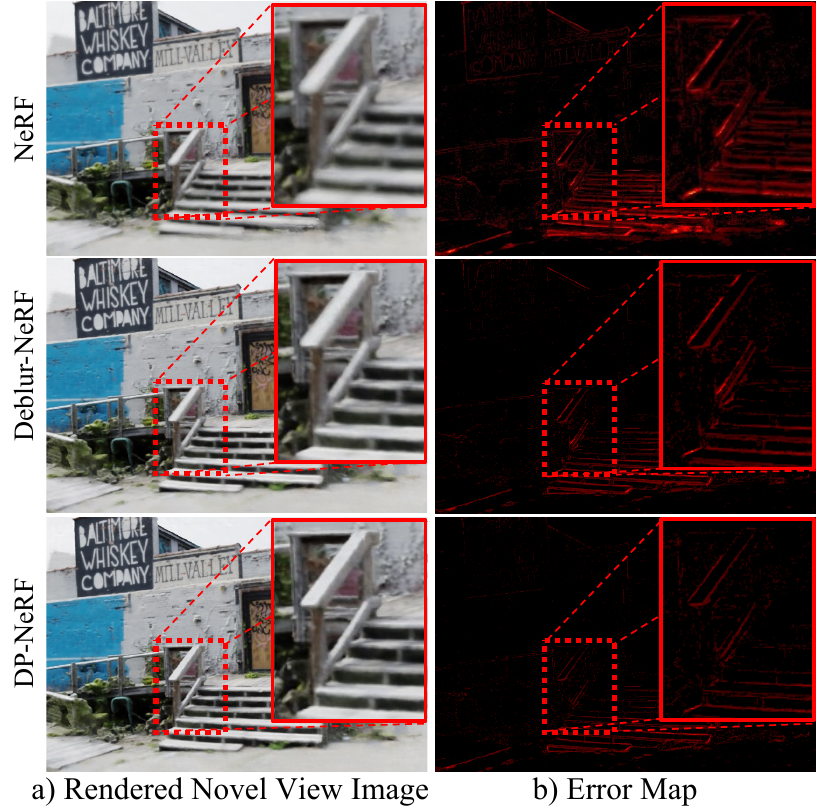} 
	  \vspace{-0.65cm}
      \caption{Visual comparison with error maps of NeRF, Deblur-NeRF, and DP-NeRF (ours) in defocus \textbf{Factory} scene. Regions with \textcolor{red}{red box} indicate emphasized regions of error map.}
      \label{fig:defocus_error_map}
      \end{center}
	  \vspace{-0.65cm}
\end{figure}

To understand the effectiveness of DP-NeRF more clearly, \figurename~\ref{fig:defocus_error_map} present an error map visualization of the stairs scene, with the brighter colors indicating greater error.
DP-NeRF produces a lower error than the baselines, reconstructing the fine details of the objects in the scene.
We also provide additional error map images without the red boxes in the supplementary material, including those for another type of blur in synthetic scenes.

\begin{figure}[t!]
      \begin{center}
         \includegraphics[scale=1.0]{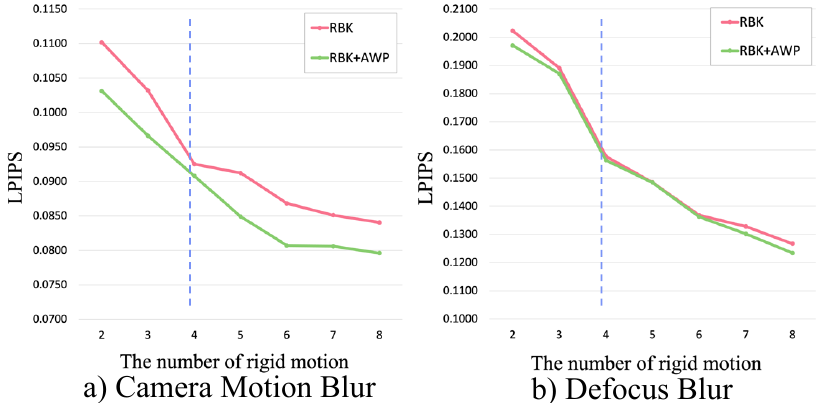} 
         \vspace{-0.6cm}
      \caption{Ablation analysis on the number of rigid camera motions for the two types of blur for synthetic \textbf{Pool} scene. (a) and (b) show the results on camera motion and defocus blur, respectively. \textcolor{magenta}{magenta} and \textcolor{yellowgreen}{yellowgreen} colors indicate results of RBK and RBK+AWP, respectively. \textcolor{blue}{blue} color line indicates the results when the number of rigid motions is $4$, which is same as of the kernel points $5$ in Deblur-NeRF~\cite{ma2022deblurnerf} for fair comparison in \tablename~\ref{tab:quantitative_results_synthetic}.}
      \label{fig:num_motion_analysis}
      \end{center}
	  \vspace{-0.95cm}
\end{figure}

\noindent \textbf{The Number of Rigid Motions.}
\figurename~\ref{fig:num_motion_analysis} presents ablation analysis for the number of rigid motions, which defines the number of transformed rays, for the two types of the blur.
 We use LPIPS as the evaluation metric because it represents perceptual quality well, as demonstrate in Nerfies~\cite{park2021nerfies}.
The results show that the performance of the RBK and RBK+AWP improves as the number of rigid motions increases, while the LPIPS for RBK+AWP is better than that for the RBK alone in all experiments.
Full quantitative PSNR, SSIM, and LPIPS results and qualitative results for the RBK and RBK+AWP are additionally presented in the supplementary material.

\noindent \textbf{RBK Analysis.}
The RBK is shown to successfully model the blur derived from both camera movement and a change in the focus plane as the rigid motion of the camera.
We demonstrate the validity of the RBK design by presenting additional kernel analysis in the supplementary material.
\vspace{-0.4cm}

\begin{figure}[t!]
      \begin{center}
         \includegraphics[scale=1.0]{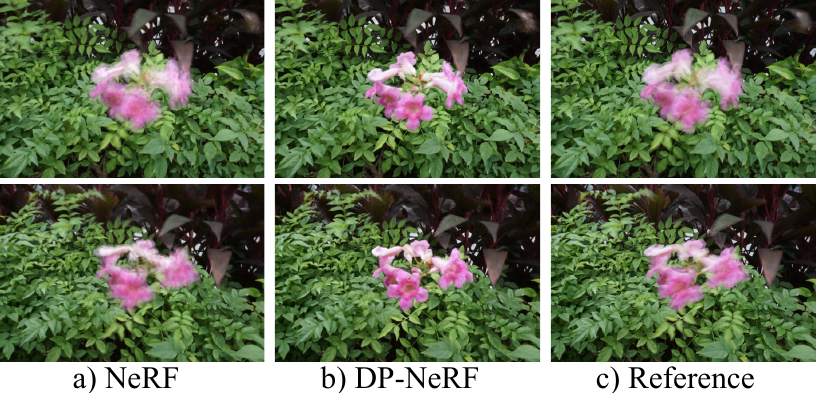} 
	  \vspace{-0.7cm}
      \caption{Visual comparison for the NeRF, the DP-NeRF, and the reference image at specific time on object motion blur. There are still in artifacts in region where motion blur occurred.}
      \label{fig:real_object_motion_failure}
      \end{center}
	  \vspace{-1.0cm}
\end{figure}
\section{Limitation \& Future Work}
\label{sec:limit}
\vspace{-0.2cm}
\noindent \textbf{Temporal Motion Blur.}
Our model fails when object motion blur is present in a scene (\figurename~\ref{fig:real_object_motion_failure}).
Though DP-NeRF produces images of higher quality than does a standard NeRF, motion blur still occurs in reconstructed scenes because we impose the physical priors based on the assumption of a static scene.
Object motion blur is an issue associated with temporal information and there is no multi-view data for a specific time in the given dataset.
This dataset structure leads to inherent geometric and appearance inconsistency in the 3D scene if we construct the NeRF without temporal modeling.
Therefore, we are confident that this limitation can be addressed with an additional temporal component in DP-NeRF in the future, such as bending the rays with $SE(3)$ field warping~\cite{park2021nerfies}, scene flow~\cite{li2021nsff}, or 3D displacement~\cite{pumarola2021dnerf} on ray samples.
Actually, although \figurename~\ref{fig:real_object_motion_failure}~(b) seems to be quite clean, the results show the explicit inconsistency when we render the spiral video to assess the 3D consistency.
Please refer the rendered videos in our project page to check the temporal inconsistency.

\noindent \textbf{Various Types of Image Noise.}
As mentioned in Section~\ref{sec:rework}, several previous studies have addressed other types of image noise for NeRF, such as temporal and exposure variation.
The modeling for these noise could be integrated with the DP-NeRF system because they are independent of each other, including our target noise.
This can be addressed in the future research if an appropriate dataset is constructed.
\vspace{-0.4cm}

\section{Conclusion}
\label{sec:cons}
\vspace{-0.2cm}
This paper proposes DP-NeRF, a novel NeRF framework from blurry inputs, that imposes the two physical priors to effectively construct a clean NeRF.
We propose the RBK to maintain geometric and appearance consistency in continuous 3D space.
In addition, We employ the AWP module to alleviate the color composition errors by considering the relationship between depth and blur.
We also introduce coarse-to-fine optimization of the two losses from the proposed modules to effectively utilize both during the training process.
Extensive experiments using synthetic and real scene datasets verified that DP-NeRF produces an improved, clean NeRF with high perceptual quality and 3D consistency in terms of geometry and appearance.
We believe that DP-NeRF represents an advance in NeRF and can be used in conjunction with other methods to construct clean NeRF, covering the images with other types of noise.

\noindent \textbf{Acknowledgements.}
This work was supported by Institute of Information \& communications Technology Planning \& Evaluation (IITP) grant funded by the Korea government(MSIT) (No.2021-0-02068, Artificial Intelligence Innovation Hub) and the Yonsei University Research Fund of 2021 (2021-22-0001).


\clearpage

\appendix
\begin{center}{\bf \Large Appendix}\end{center}\vspace{-2mm}
\section{Additional Implementation Details}
\label{sec:more_imple_detail}
\subsection{Training}
\label{subsec:imple_training}
DP-NeRF is implemented on two Nvidia RTX 3090 GPUs based on the published code and dataset for Deblur-NeRF~\cite{ma2022deblurnerf} using PyTorch~\cite{paszke2017pytorch}.
Training images are resized to $600\times400$ across the entire dataset to train the DP-NeRF.
In addition, we start to optimize proposed components, rigid blurring kernel (RBK), adaptive weight proposal (AWP), and color composition (CC), after 1200 training iterations with a pure NeRF~\cite{mildenhall2020nerf} to obtain coarse scene representation.

\subsection{Architectural Detail}
\label{subsec:imple_architecture}
\begin{figure}[t]
      \begin{center}
         \includegraphics[scale=1.0]{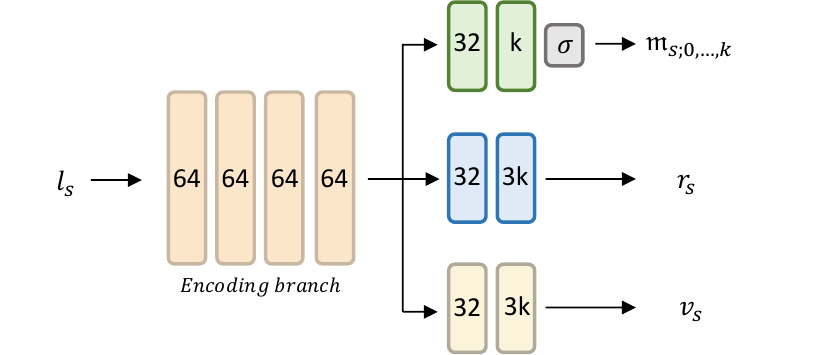} 
      \caption{A detailed architecture of rigid blurring kernel (RBK).}
      \label{fig:rbk_architecture}
      \end{center}
	  \vspace{-0.5cm}
\end{figure}

\begin{figure}[t]
      \begin{center}
         \includegraphics[scale=1.0]{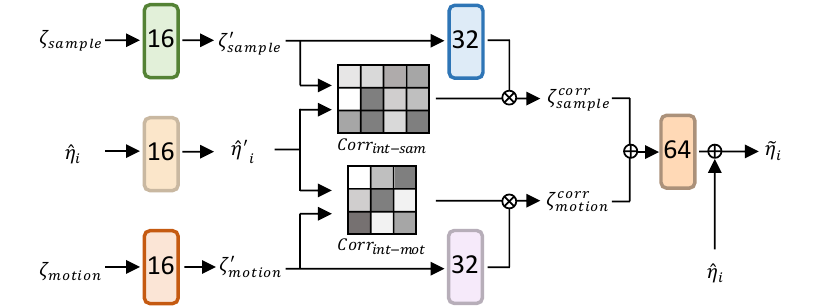} 
      \caption{Detailed description about a correlation part of motion aggregation module (MAM). This picture includes process for computing correlation between $\zeta^{p}_{motion}$, $\zeta^{p}_{sample}$, and $\hat{\eta}^{p}_{i}$.}
      \label{fig:mam_correlation}
      \end{center}
	  \vspace{-0.4cm}
\end{figure}

Figures~~\ref{fig:rbk_architecture}, \ref{fig:mam_correlation}, \ref{fig:DP-NeRF_pipeline}, and \ref{fig:mam_architecture} describe the DP-NeRF architecture to describe in detail.
View information $l_{s}$ for each image is embedded with 64 channels in a simple embedding layer.

\noindent \textbf{Rigid Blurring Kernel (RBK).} 
As shown in \figurename~\ref{fig:rbk_architecture}, RBK consists of one shared encoding branch $E$ and three decoding branches with simple MLPs ($\mathcal{W}$, $\mathcal{R}$, and $\mathfrak{L}$).
Encoding branch $E$ consists of an MLP with four fully-connected linear layers, with each layer having 64 dimensions and ReLU activation function.
Decoding branches for $\mathfrak{m}_{s;0,\dots,k}$, $r_{s}$, and $v_{s}$ also consist of an MLP with one linear layer with 32 dimensions and an output linear layer.
The output channel dimensions for $\mathfrak{m}_{s;0,\dots,k}$, $r_{s}$, and $v_{s}$ is $k$, $3k$, and $3k$, respectively, where $k$ is a hyper-parameter that controls the number of rigid camera motions.
Note that, $\mathfrak{m}_{s;0,\dots,k}$ is normalized along the motion axis $k$ to ensure $\sum^{k}_{i=0}\mathfrak{m}_{s;i}=1$.
\vspace{-0.2cm}

\noindent \textbf{Adaptive Weight Proposal (AWP).} 
First of all, we additionally describe a motivation of the complex architecture design of AWP.
Intuition of AWP architecture is to fully use the spatial occupancy information of samples on the rays. 
When we use the similar module based on inter-motion correlation with rendered depth values on each rays, we could not get the improved results.
The reason was supposed to be insufficient occupancy information of one scalar rendered depth value per ray.
Hence, we design AWP to fully reflect the information to use rich correlation between the rays.
Here are description of AWP architecture with detailed notations.
Please refer to \figurename~\ref{fig:mam_correlation}, \figurename~\ref{fig:mam_architecture}, and \tablename~\ref{tab:awp_notation} for architecture design and corresponding  notations.
Note that, we omit the notation for the batch dimension for clarity.
Before forwarding to motion aggregation module (MAM), extracted depth feature $\zeta^{p}_{i,j}\in\mathbb{R}^{N_{m}\times N_{s}\times C_{d1}}$ from the second-to-last layer of the NeRF is embedded in $\hat{\zeta}^{p}_{i,j}\in\mathbb{R}^{N_{m}\times N_{s}\times C_{d2}}$ via the simple four-layered MLP with ReLU activation, where $N_{m}$ and $N_{s}$ denote the dimensions of the motion axis and sample axis, respectively. 
$N_{m}$ is the number of blurring rays, which is a summation of the number of motion ($k$) and original ray ($1$).
$N_{s}$ is the total number of samples, which is a summation of the number of coarse samples ($N_{c}$) and fine samples ($N_{f}$).
We then, apply feature modulation (FM)~\cite{zhao2022nerfattention} to generate ray-wise representative features, $\eta^{p}_{i}\in \mathbb{R}^{N_{m}\times C_{d2}}$.
To impose the view- and direction-information for the rays, $l_{s}$ and positional embedded $\textbf{d}^{p-rig}_{s;0,\dots,k}$ are concatenated and forwarded to the simple MLP to extract $\hat{\eta}^{p}_{i}\in\mathbb{R}^{N_{m}\times C_{d3}}$.
To aggregate the implicit information between the modeled blurring rays, we forward $\hat{\zeta}^{p}_{i,j}$ and $\hat{\eta}^{p}_{i}$ to the MAM.
The MAM then aggregates the extracted features from $\hat{\zeta}^{p}_{i,j}$ and $\hat{\eta}^{p}_{i}$ based on the computed correlation between the decomposed information along each motion and sample axis.


\begin{table}[t]
   \tiny
   \begin{center}
      \resizebox{\columnwidth}{!}{
		\setlength{\tabcolsep}{1pt}
         \begin{tabular}{c||c|c|c|c|c|c|c|c|c|c}
         	\toprule
         	~Notation~&~$l_{s}$~&~$C_{d1}$~&~$C_{d2}$~&~$C_{d3}$~&~$C_{d4}$~&~$N_{m}$~&~$N_{s}$~&~$N_{c}$~&~$N_{f}$~\\
			\midrule
			Value & 64 & 128 & 64 & 32 & 16 & ($1+k$) & ($N_{c}+N_{f}$) & 64 & 64 \\
			\bottomrule
      \end{tabular}
      }
   \vspace{-0.2cm}
   \caption{Notations for adaptive weight proposal (AWP) and corresponding values used in our experiment as default.}
   \label{tab:awp_notation}
   \end{center}
   \vspace{-0.8cm}
\end{table}

\begin{figure*}[t]
      \captionsetup{justification=centering}
		\includegraphics[width=2.1\columnwidth]{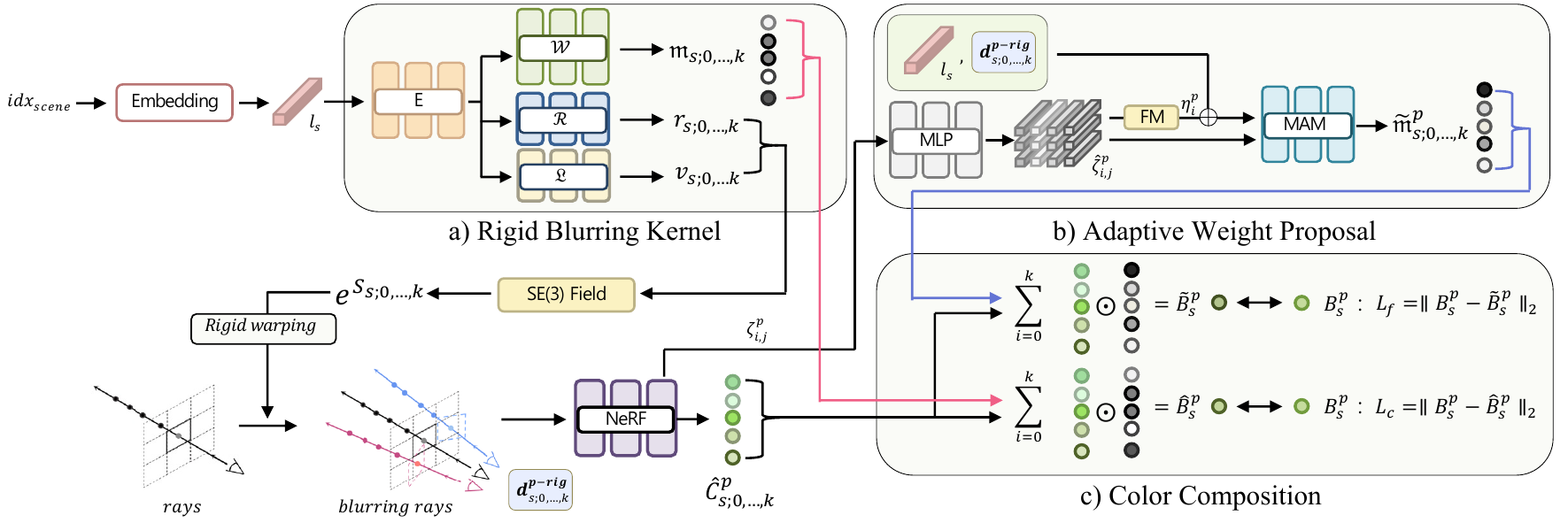}
      \vspace{-0.6cm}
      \caption{Overall pipeline of DP-NeRF, which is same as figure in the main paper.}
      \vspace{-0.4cm}
      \label{fig:DP-NeRF_pipeline}
   \end{figure*}

The MAM is formulated in the main paper as follows:
\begin{equation} \label{eq:mam_detail}
MAM(\hat{\zeta}^{p}_{i;j}, \hat{\eta}^{p}_{i})=MLP\big(cat(\textbf{corr}(\hat{\eta}^{p}_{i}, \xi^{p}_{motion}, \xi^{p}_{sample}))\big),
\end{equation}
where $\zeta^{p}_{motion}\in \mathbb{R}^{N_{s}\times C_{d3}}$ and $\zeta^{p}_{sample}\in \mathbb{R}^{N_{m}\times C_{d3}}$ are embedded via a one-layer MLP and attentive pooling~\cite{hu2020attentivepooling} along the motion and sample axis.
Modulated feature $\eta^{p}_{i}$ and attentive pooled features $\zeta^{p}_{motion}$ and $\zeta^{p}_{sample}$ are embedded again in $\hat{\eta}'^{p}_{i}$, $\zeta'^{p}_{sample}$ and $\zeta'^{p}_{motion}$ as the same channel dimension $C_{d4}$ before computing the correlation maps.
We then compute two correlations, $Corr_{int-mot}$ and $Corr_{int-sam}$, which represent the inter-motion and inter-sample correlations, respectively (\figurename~\ref{fig:mam_correlation}).
Because $\hat{\zeta}^{p}_{i,j}$ consists of a sample dimension and a motion dimension, we apply inter-sample and inter-motion design to investigate the inter-sample and inter-motion correlations.
The each type of correlation score maps enhances the correlation of each dimension and aggregate them with modulated features via matrix multiplication.
The enhanced features, $\zeta^{corr}_{sample}$ and $\zeta^{corr}_{motion}$, are then concatenated and forwarded to the 64-channel MLP, followed by the residual connection of $\hat{\eta}_{i}$ to generate our final motion-aggregated feature $\tilde{\eta}_{i}$. Note that, the dimensions of  $\hat{\eta}_{i}$ and $\tilde{\eta}_{i}$ are the same.

\noindent \textbf{Tone Mapping.} 
Before color composition for each color from the blurring rays and the composition weights $\mathfrak{m}_{s;0,\dots,k}$ or $\tilde{\mathfrak{m}}^{p}_{s;0,\dots,k}$, we should note that each color $\hat{C}^{p}_{s;0,\dots,k}$ is tone-mapped from the predicted scene irradiance from a NeRF similar to \cite{ma2022deblurnerf}.
Gamma function $g$ for the tone mapping function is simply set as shown in Eq.~\ref{eq:gamma_function} in a similar manner to \cite{ma2022deblurnerf} because there is no significant difference in performances when either a gamma function or learnable MLP is employed as the tone mapping function.
\vspace{-0.1cm}
\begin{equation} \label{eq:gamma_function}
g(c')=c'^{\frac{1}{2.2}},
\vspace{-0.1cm}
\end{equation}
where $c'$ denotes the predicted radiance from the NeRF in DP-NeRF.
The types of tone-mapping function does not significantly influence the predicted radiance due to the consistent exposure of the dataset.
Hence, we can rewrite the full color composition of $\hat{B}^{p}_{s}$ and $\tilde{B}^{p}_{s}$ as shown in Eq.~\ref{eq:full_color_composition}, with $g$ omitted in the main paper due to page limitations.
\vspace{-0.1cm}
\begin{align} \label{eq:full_color_composition}
\begin{split}
\hat{B}^{p}_{s} & = \mathfrak{m}_{s;0}~g(\hat{C}^{p}_{s;0}) + \sum_{i=1}^{k}\mathfrak{m}_{s;i}~g(\hat{C}^{p-rig}_{s;i})\\
\tilde{B}^{p}_{s} & = \tilde{\mathfrak{m}}^{p}_{s;0}~g(\hat{C}^{p}_{s;0}) + \sum_{i=1}^{k}\tilde{\mathfrak{m}}^{p}_{s;i}~g(\hat{C}^{p-rig}_{s;i})
\end{split}
\end{align}
\vspace{-0.6cm}

\begin{figure}[t]
      \begin{center}
         \includegraphics[scale=1.0]{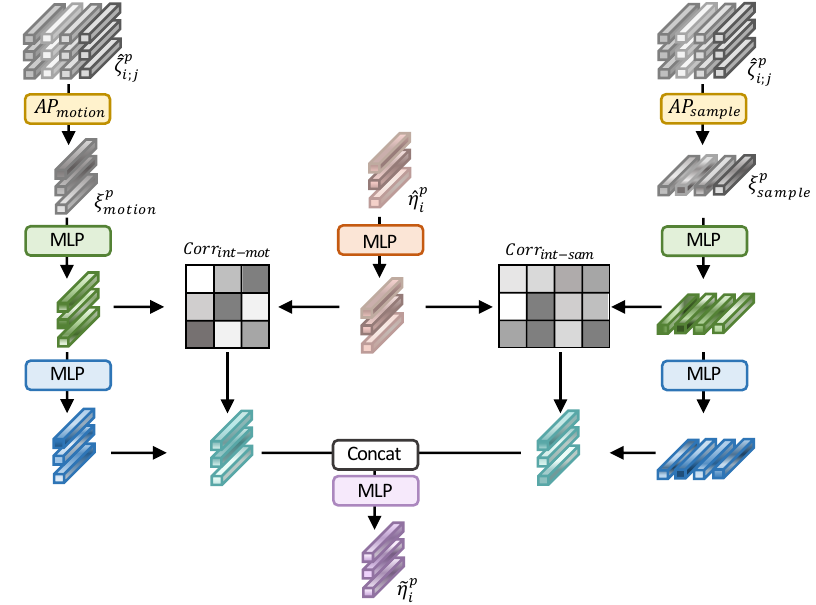} 
      \caption{Architecture of motion aggregation module (MAM), which is same as figure in the main paper.}
      \label{fig:mam_architecture}
      \end{center}
	  \vspace{-0.8cm}
\end{figure}

\section{Evaluation Results on Real Scene}
\label{sec:eval_real}
\noindent \textbf{Quantitative Results.} 
We present quantitative evaluation results for camera motion and defocus blur in the real scene dataset in Tables~\ref{tab:quantitative_results_real_camera_motion} and \ref{tab:quantitative_results_real_defocus}.
The results show that DP-NeRF improves the quantitative performance for all of the evaluation metrics, especially on LPIPS.
In addition, it is clear that PSNR and SSIM cannot fully represent the realistic perceptual quality of rendered images as argued by Nerfies~\cite{park2021nerfies} in their paper.
Hence, the perceptual quality should be evaluated by comparing LPIPS and the rendered image quality visually.

\noindent \textbf{Qualitative Results.} 
We present the qualitative results in Figures~\ref{fig:qualitative_results_real_camera_motion} and \ref{fig:qualitative_results_real_defocus}.
The results show that DP-NeRF(RBK) and DP-NeRF(RBK+AWP) both outperform the NeRF and baseline in terms of perceptual quality for most of the scenes.
It demonstrates that our model produces more accurate 3D reconstruction quality with a realistic, clean NeRF.
In addition, DP-NeRF enhances the 3D geometric and appearance consistency in several scenes.
For example, due to the complex depth and similar texture of the forest region, baselines~\cite{mildenhall2020nerf,ma2022deblurnerf} have difficulty in predicting the correct geometry in the \textbf{Heron} scene (4th-row in \figurename~\ref{fig:qualitative_results_real_camera_motion}).
However, our model predicts the region more accurately, illustrating that DP-NeRF can model complex 3D space more accurately.

\section{Additional Ablation Results}
\label{sec:more_abls}
\subsection{RBK Analysis}
\label{subsec:rbk_analysis}
\begin{figure*}[ht]
      \centering
         \includegraphics[scale=1.1]{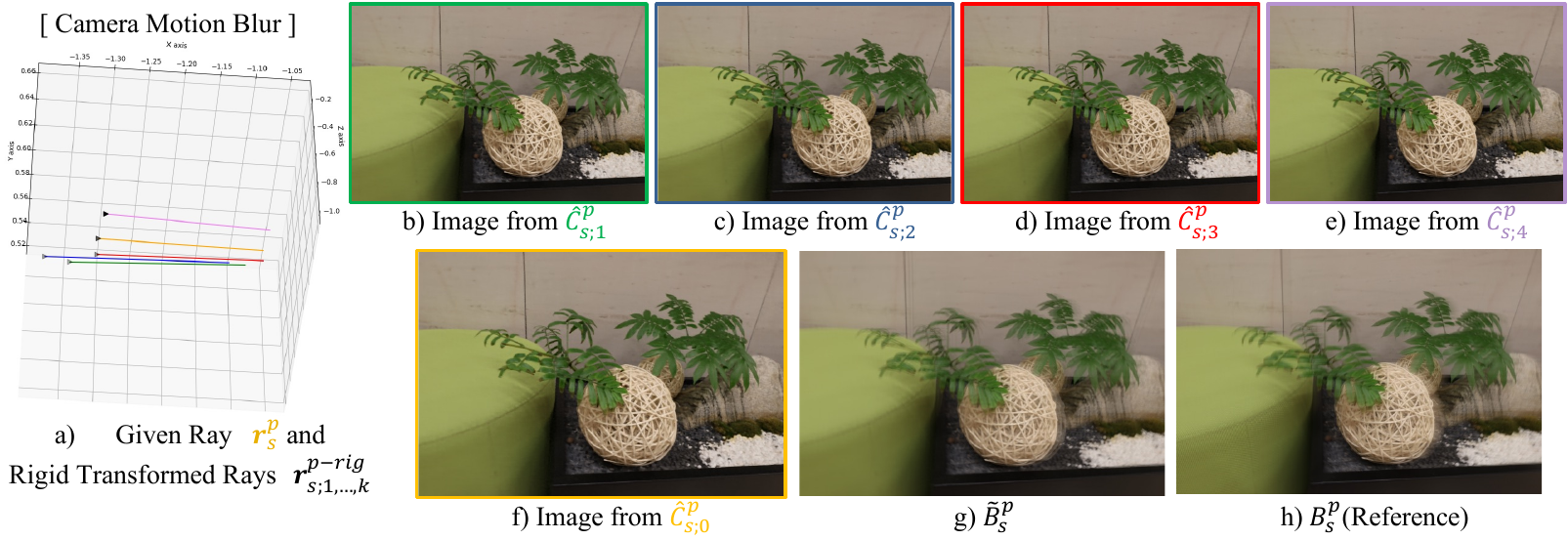} 
      \caption{RBK analysis on camera motion blur for real \textbf{blurball} scene. \figurename~(a) denotes visualization result of original and rigid transformed camera origin and direction on given reference view, which is presented in \figurename~(h). Figures (b)-(f) denote rendered images from given and transformed cameras in \figurename~(a). \figurename~(g) denotes a composited blurred image from Figures (b)-to-(f) with composition weights to predict a reference image, \figurename~(h). Figures (g) and (h) are used in DP-NeRF as RGB reconstruction loss.}
      \label{fig:rbk_analysis_cameramotion}
   \end{figure*}

\begin{figure*}[ht]
      \centering
         \includegraphics[scale=1.1]{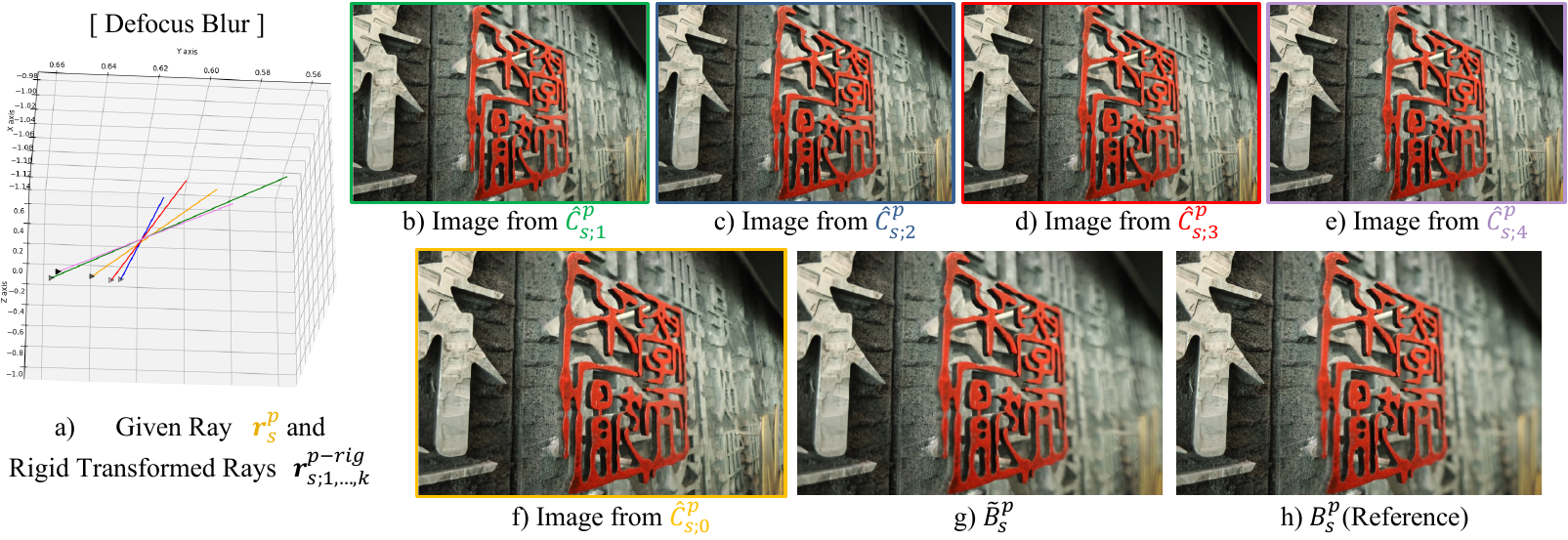} 
      \caption{RBK analysis on defocus blur for real \textbf{defocusseal} scene. \figurename~(a) denotes visualization result of original and rigid transformed camera origin and direction on given reference view, which is presented in \figurename~(h). Figures (b)-(f) denote rendered images from camera and transformed cameras in \figurename~(a). \figurename~(g) denotes a composited blurred image from Figures (b)-(f) with composition weights to predict a reference image, \figurename~(h). \figurename~(g) and (h) are used in DP-NeRF as RGB reconstruction loss.}
      \label{fig:rbk_analysis_defocus}
   \end{figure*}

\noindent \textbf{Modeling Analysis.}
We illustrate how our RBK models camera motion and defocus blur as ray rigid transformation by presenting a visualization of the modeled kernel and rendered images from each of the transformed camera views in Figures~\ref{fig:rbk_analysis_cameramotion} and \ref{fig:rbk_analysis_defocus}.
Figures~\ref{fig:rbk_analysis_cameramotion}~(a) and \ref{fig:rbk_analysis_defocus}~(a) shows the transformed ray origin and direction derived from the trained DP-NeRF with paired images.
Note that, each colored transformed camera origin and ray direction pairs with a rendered image with the same colored box.
Each image also has the same notation as presented in the main paper~($\hat{C}^{p}_{s;0}$ and $\hat{C}^{p}_{s;1,\dots,k}$) to assist in understanding.
Figures~\ref{fig:rbk_analysis_cameramotion}~(b)-(f) and \ref{fig:rbk_analysis_defocus}~(b)-(f) present rendered images from the transformed cameras, which are used to composite the blurred image $\tilde{B}^{p}_{s}$.
\figurename~\ref{fig:rbk_analysis_cameramotion}~(g) and \ref{fig:rbk_analysis_defocus}~(g) present the composited blurred image from images (b)-(f) with composition weights as we mentioned in the main paper.
We demonstrate that the composited blurred image $\tilde{B}^{p}_{s}$ is successfully generated, with a similar appearance to the reference image $B^{p}_{s}$~(Figures~\ref{fig:rbk_analysis_cameramotion}~(h) and \ref{fig:rbk_analysis_defocus}~(h)).
Figures~\ref{fig:rbk_analysis_cameramotion} and \ref{fig:rbk_analysis_defocus} show that DP-NeRF can model the blurred image $\tilde{B}^{p}_{s}$ so that it is similar to reference image $B^{p}_{s}$ for both type of the blur.

The visualizations of given ray $\textbf{r}^{p}_{s}(=\textbf{r}^{p}_{s;0}) $ and rigidly transformed~(RT) rays $\textbf{r}^{p-rig}_{s;1,\dots,k}$ in Figures \ref{fig:rbk_analysis_cameramotion}~(a) and \ref{fig:rbk_analysis_defocus}~(a) demonstrate that the camera motion and defocus blur can be successfully modeled with the RBK imitating camera movement or focus plain decision by rigidly warping the given scene camera with the $SE(3)$ field.
It is obvious that camera motion blur can be successfully modeled using the RBK because it models the blurring process using rigid camera transformation.
\figurename~\ref{fig:rbk_analysis_cameramotion}~(a) presents the results of the camera shaking during image acquisition, which leads to camera motion blur.

Interestingly, for defocus blur, there is a plane where the RT rays intersect  (\figurename~\ref{fig:rbk_analysis_defocus}~(a)), and this is the predicted focus plane.
Specifically, RBK imitates the defocus blurring by approximating the depth of field process locating the transformed rays on the virtual aperture. The predicted transforms of the rays naturally decide the focus plane as we described in \figurename~\ref{fig:rbk_analysis_defocus}~(a).
If the depth value of a given ray is not on focus plane, focus blur is naturally induced by the color composition of the given ray and the other RT rays, which penetrate the other surrounding parts of the scene.

In addition, Figures~\ref{fig:rbk_analysis_cameramotion}~(b)-(h) and \ref{fig:rbk_analysis_defocus}~(b)-(h) show that we can render images related to the construction of the blurred image for a scene.
This form of image decomposition can determine how the blurred image in a scene is generated during image acquisition with respect to change in the camera settings, such as camera motion or focus plain information.
Furthermore, RBK modeling guarantees the geometric and appearance consistency of each rendered image.

\subsection{Effectiveness of the RBK and AWP}
\label{subsec:rbk_awp}
\begin{figure*}[t]
      \centering
         \includegraphics[scale=1.0]{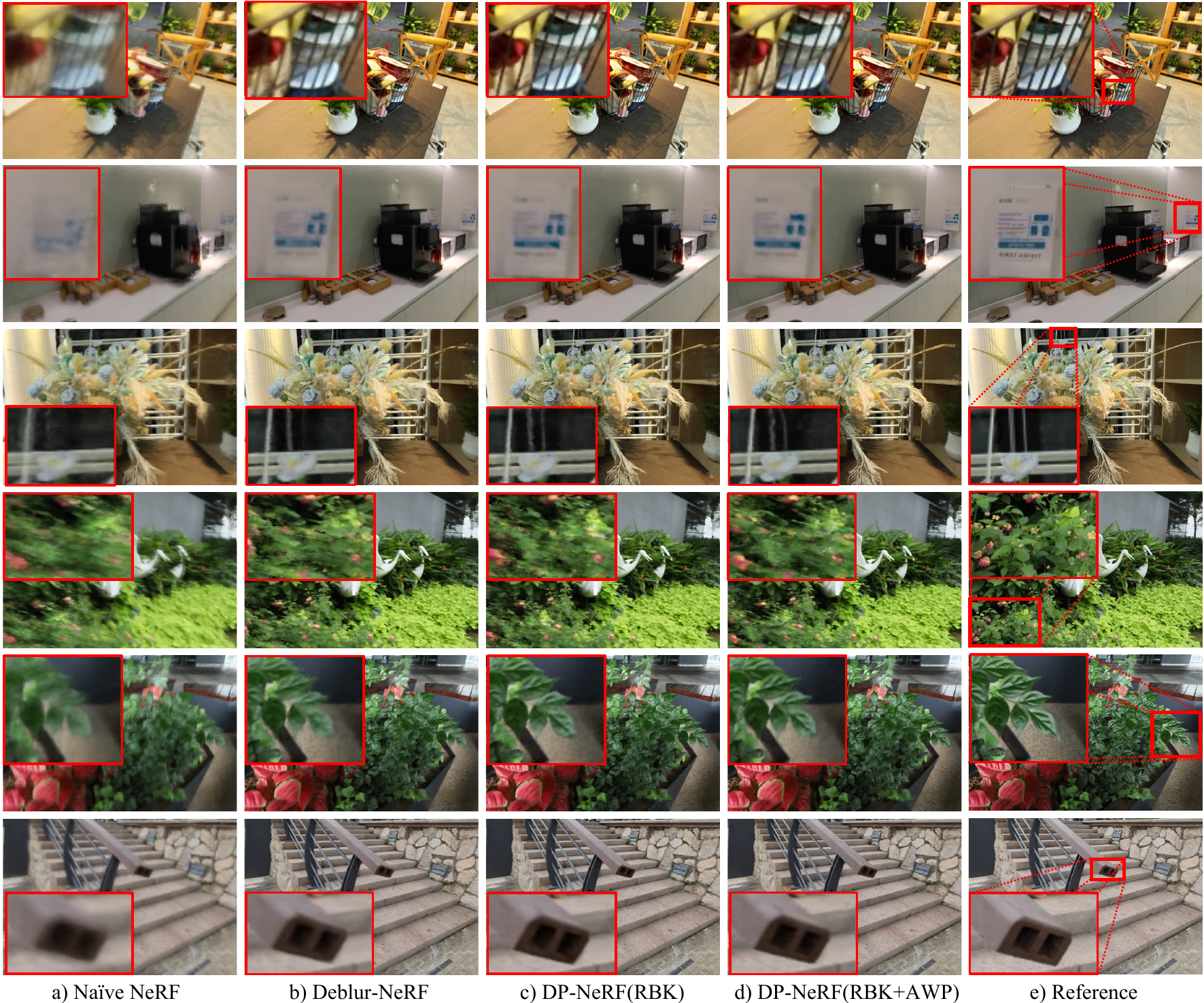} 
      \caption{Rendered novel view synthesis results of DP-NeRF for real scene camera motion blur. Figures (a)-(e) denote Naive NeRF, Deblur-NeRF, ours(RBK), ours(RBK+AWP), and ground truth images, respectively. \textcolor{red}{Red} colored box in corner of images are enlarged part of same colored box region in reference images.}
      \label{fig:qualitative_results_real_camera_motion}
   \end{figure*}

\begin{figure*}[t]
         \includegraphics[scale=1.0]{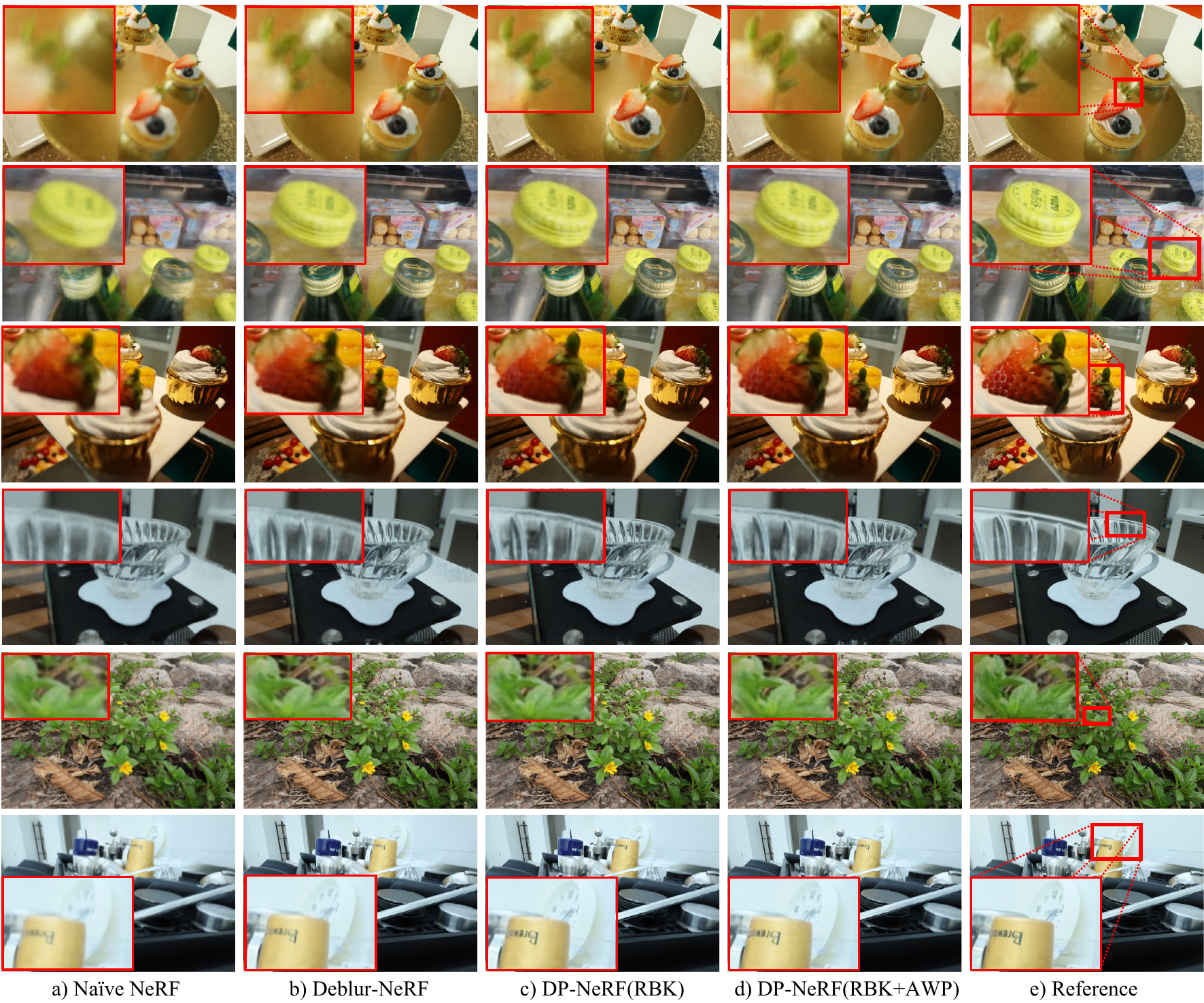} 
      \caption{Rendered novel view synthesis results of DP-NeRF for real scene defocus blur. Figures (a)-(e) denote Naive NeRF, Deblur-NeRF, ours(RBK), ours(RBK+AWP), and ground truth images, respectively. \textcolor{red}{Red} colored box in corner of images are enlarged part of same colored box region in reference images.}
      \label{fig:qualitative_results_real_defocus}
   \end{figure*}

We present the effectiveness of the RBK and AWP with ablation analysis using the synthetic~(\tablename~\ref{tab:ablation_synthetic}) and real scene~(Tables~\ref{tab:ablation_real_camera_motion} and \ref{tab:ablation_real_defocus} for camera motion and defocus blur, respectively) datasets.
Qualitative comparisons are also presented in Figures~\ref{fig:qualitative_results_real_camera_motion} and \ref{fig:qualitative_results_real_defocus}.
For real scene dataset, it seems to be marginal improvement with AWP. 
Actually, blurring kernel is affected by the depth in case of the out-of-plane camera motion blur and general defocus blur as mentioned in \cite{srinivasan2017light,srinivasan2018aperture}. 
However, for provided camera motion blur dataset, the blur type is close to in-plane camera motion blur, which leads to the marginal improvement.
In addition, shape of the blurring kernel can change depending the depth in both blur types.
Instead to directly model kernel shape change, we mitigate the issue through adaptive weights between transformed rays from AWP. 
Our model shows the more satisfying results with geometrically clean images with RBK. 
In addition, we can find that our model with AWP get more detailed clean results thanks to above adaptive weights approximation. 

\clearpage

\begin{table*}[t]
   \Large
   \begin{center}
      \resizebox{2.1\columnwidth}{!}{
		\centering
		\setlength{\tabcolsep}{1pt}
         \begin{tabular}{c||ccc|ccc|ccc|ccc|ccc|ccc}
	        \specialrule{0.6pt}{1pt}{1pt}
			\multirow{2}{*}{Camera Motion}& \multicolumn{3}{c}{Ball} & \multicolumn{3}{c}{Basket} & \multicolumn{3}{c}{Buick} & \multicolumn{3}{c}{Coffee} & \multicolumn{3}{c}{Decoration} & & \\
			 & PSNR($\uparrow$) & SSIM($\uparrow$) & LPIPS($\downarrow$) & PSNR($\uparrow$) & SSIM($\uparrow$) & LPIPS($\downarrow$) & PSNR($\uparrow$) & SSIM($\uparrow$) & LPIPS($\downarrow$) & PSNR($\uparrow$) & SSIM($\uparrow$) & LPIPS($\downarrow$) & PSNR($\uparrow$) & SSIM($\uparrow$) & LPIPS($\downarrow$) &  & &\\
			\midrule
			Naive NeRF~\cite{mildenhall2020nerf} & 24.08 & 0.6237 & .3992 & 23.72 & .7086 & .3223 & 21.59 & .6325 & .3502 & 26.48 & .8064 & .2896 & 22.39 & .6609 & .3633 & & & \\
			Deblur-NeRF~\cite{ma2022deblurnerf} & \cellcolor{orange!25}27.36 & \cellcolor{orange!25}.7656 & \cellcolor{yellow!25}.2230 & \cellcolor{yellow!25}27.67 & \cellcolor{yellow!25}.8449 & \cellcolor{yellow!25}.1481 & \cellcolor{yellow!25}24.77 & \cellcolor{yellow!25}.7700 & \cellcolor{yellow!25}.1752 & \cellcolor{yellow!25}30.93 & \cellcolor{yellow!25}.8981 & \cellcolor{yellow!25}.1244 & \cellcolor{yellow!25}24.19 & \cellcolor{yellow!25}.7707 & \cellcolor{yellow!25}.1862 & & & \\
			DP-NeRF & \cellcolor{yellow!25}27.20 & \cellcolor{yellow!25}.7652 & \cellcolor{orange!25}.2088 & \cellcolor{orange!25}27.74 & \cellcolor{orange!25}.8455 & \cellcolor{orange!25}.1294 & \cellcolor{orange!25}25.70 & \cellcolor{orange!25}.7922 & \cellcolor{orange!25}.1405 & \cellcolor{orange!25}31.19 & \cellcolor{orange!25}.9049 & \cellcolor{orange!25}.1002 & \cellcolor{orange!25}24.31 & \cellcolor{orange!25}.7811 & \cellcolor{orange!25}.1639 & & & \\
			\midrule\midrule
			\multirow{2}{*}{Camera motion}& \multicolumn{3}{c}{Girl} & \multicolumn{3}{c}{Heron} & \multicolumn{3}{c}{Parterre} & \multicolumn{3}{c}{Puppet} & \multicolumn{3}{c}{Stair} & \multicolumn{3}{c}{Average} \\
			 & PSNR($\uparrow$) & SSIM($\uparrow$) & LPIPS($\downarrow$) & PSNR($\uparrow$) & SSIM($\uparrow$) & LPIPS($\downarrow$) & PSNR($\uparrow$) & SSIM($\uparrow$) & LPIPS($\downarrow$) & PSNR($\uparrow$) & SSIM($\uparrow$) & LPIPS($\downarrow$) & PSNR($\uparrow$) & SSIM($\uparrow$) & LPIPS($\downarrow$) & PSNR($\uparrow$) & SSIM($\uparrow$) & LPIPS($\downarrow$) \\
			\midrule
			Naive NeRF~\cite{mildenhall2020nerf} & 20.07 & .7075 & .3196 & 20.50 & .5217 & .4129 & 23.14 & .6201 & .4046 & 22.09 & .6093 & .3389 & 22.87 & .4561 & .4868 & 22.69 & .6347 & .3687 \\
			Deblur-NeRF~\cite{ma2022deblurnerf} & \cellcolor{yellow!25}22.27 & \cellcolor{yellow!25}.7976 & \cellcolor{yellow!25}.1687 & \cellcolor{yellow!25}22.63 & \cellcolor{yellow!25}.6874 & \cellcolor{yellow!25}.2099 & \cellcolor{yellow!25}25.82 & \cellcolor{yellow!25}.7597 & \cellcolor{yellow!25}.2161 & \cellcolor{yellow!25}25.24 & \cellcolor{yellow!25}.7510 & \cellcolor{yellow!25}.1577 & \cellcolor{yellow!25}25.39 & \cellcolor{yellow!25}.6296 & \cellcolor{yellow!25}.2102 & \cellcolor{yellow!25}25.63 & \cellcolor{yellow!25}.7675& \cellcolor{yellow!25}.1820 \\
			DP-NeRF & \cellcolor{orange!25}23.33 & \cellcolor{orange!25}.8139 & \cellcolor{orange!25}.1498 & \cellcolor{orange!25}22.88 & \cellcolor{orange!25}.6930 & \cellcolor{orange!25}.1914 & \cellcolor{orange!25}25.86 & \cellcolor{orange!25}.7665 & \cellcolor{orange!25}.1900 & \cellcolor{orange!25}25.25 & \cellcolor{orange!25}.7536 & \cellcolor{orange!25}.1505 & \cellcolor{orange!25}25.59 & \cellcolor{orange!25}.6349 & \cellcolor{orange!25}.1772 & \cellcolor{orange!25}25.91 & \cellcolor{orange!25}.7751 & \cellcolor{orange!25}.1602 \\
	        \specialrule{0.6pt}{1pt}{1pt}
			\bottomrule
      \end{tabular}
      }%
   \caption{Quantitative results for the real scene camera motion blur. Each color shading indicates the \colorbox{orange!25}{best} and \colorbox{yellow!25}{second-best} result, respectively.}
   \label{tab:quantitative_results_real_camera_motion}
   \end{center}
\end{table*}
\clearpage

\begin{table*}[t]
   \Large
   \begin{center}
      \resizebox{2.1\columnwidth}{!}{
		\centering
		\setlength{\tabcolsep}{1pt}
         \begin{tabular}{c||ccc|ccc|ccc|ccc|ccc|ccc}
	        \specialrule{0.6pt}{1pt}{1pt}
			\multirow{2}{*}{Defocus} & \multicolumn{3}{c}{Cake} & \multicolumn{3}{c}{Caps} & \multicolumn{3}{c}{Cisco} & \multicolumn{3}{c}{Cupcake} & \multicolumn{3}{c}{Coral} & \multicolumn{3}{c}{} \\
			 & PSNR($\uparrow$) & SSIM($\uparrow$) & LPIPS($\downarrow$) & PSNR($\uparrow$) & SSIM($\uparrow$) & LPIPS($\downarrow$) & PSNR($\uparrow$) & SSIM($\uparrow$) & LPIPS($\downarrow$) & PSNR($\uparrow$) & SSIM($\uparrow$) & LPIPS($\downarrow$) & PSNR($\uparrow$) & SSIM($\uparrow$) & LPIPS($\downarrow$) & & & \\
			\midrule
			Naive NeRF & 24.42 & .7210 & .2250 & 22.73 & .6312 & .2801  & 20.72 & .7217 & .1256 & 21.88 & .6809 & .2155 & 19.81 & .5658 & .2689 & & &  \\
			Deblur-NeRF~\cite{ma2022deblurnerf} & \cellcolor{orange!25}26.27 & \cellcolor{orange!25}.7800 & \cellcolor{yellow!25}.1282 & \cellcolor{yellow!25}23.87 & \cellcolor{orange!25}.7128 & \cellcolor{yellow!25}.1612 & \cellcolor{orange!25}20.83 & \cellcolor{orange!25}.7270 & \cellcolor{yellow!25}.0868 & \cellcolor{yellow!25}22.26 & \cellcolor{yellow!25}.7219 & \cellcolor{yellow!25}.1160 & \cellcolor{yellow!25}19.85 & \cellcolor{yellow!25}.5999 & \cellcolor{yellow!25}.1214 & & & \\
			DP-NeRF & \cellcolor{yellow!25}26.16 & \cellcolor{yellow!25}.7781 & \cellcolor{orange!25}.1267 & \cellcolor{orange!25}23.95 & \cellcolor{yellow!25}.7122 & \cellcolor{orange!25}.1430 & \cellcolor{yellow!25}20.73 & \cellcolor{yellow!25}.7260 & \cellcolor{orange!25}.0840 & \cellcolor{orange!25}22.80 & \cellcolor{orange!25}.7409 & \cellcolor{orange!25}.0960 & \cellcolor{orange!25}20.11 & \cellcolor{orange!25}.6107 & \cellcolor{orange!25}.1178 & & & \\

			\midrule\midrule
			\multirow{2}{*}{Defocus}& \multicolumn{3}{c}{Cups} & \multicolumn{3}{c}{Daisy} & \multicolumn{3}{c}{Sausage} & \multicolumn{3}{c}{Seal} & \multicolumn{3}{c}{Tools} & \multicolumn{3}{c}{Average} \\
			 & PSNR($\uparrow$) & SSIM($\uparrow$) & LPIPS($\downarrow$) & PSNR($\uparrow$) & SSIM($\uparrow$) & LPIPS($\downarrow$) & PSNR($\uparrow$) & SSIM($\uparrow$) & LPIPS($\downarrow$) & PSNR($\uparrow$) & SSIM($\uparrow$) & LPIPS($\downarrow$) & PSNR($\uparrow$) & SSIM($\uparrow$) & LPIPS($\downarrow$) & PSNR($\uparrow$) & SSIM($\uparrow$) & LPIPS($\downarrow$) \\
			\midrule
			Naive NeRF~\cite{mildenhall2020nerf} & 25.02 & .7581 & .2315 & 22.74 & .6203 & .2621 & 17.79 & .4830 & .2789 & 22.79 & .6267 & .2680 & 26.08 & .8523 & .1547 & 22.40 & .6661 & .2310\\
			Deblur-NeRF~\cite{ma2022deblurnerf} & \cellcolor{yellow!25}26.21 & \cellcolor{yellow!25}.7987 & \cellcolor{yellow!25}.1271 & \cellcolor{yellow!25}23.52 & \cellcolor{yellow!25}.6870 & \cellcolor{yellow!25}.1208 & \cellcolor{yellow!25}18.01 & \cellcolor{yellow!25}.4998 & \cellcolor{yellow!25}.1796 & \cellcolor{orange!25}26.04 & \cellcolor{yellow!25}.7773 & \cellcolor{yellow!25}.1048 & \cellcolor{yellow!25}27.81 & \cellcolor{yellow!25}.8949 & \cellcolor{yellow!25}.0610 & \cellcolor{yellow!25}23.46 & \cellcolor{yellow!25}.7199 & \cellcolor{yellow!25}.1207\\
			DP-NeRF & \cellcolor{orange!25}26.75 & \cellcolor{orange!25}.8136 & \cellcolor{orange!25}.1035 & \cellcolor{orange!25}23.79 & \cellcolor{orange!25}.6971 & \cellcolor{orange!25}.1075 & \cellcolor{orange!25}18.35 & \cellcolor{orange!25}.5443 & \cellcolor{orange!25}.1473 & \cellcolor{yellow!25}25.95 & \cellcolor{orange!25}.7779 & \cellcolor{orange!25}.1026 & \cellcolor{orange!25}28.07 & \cellcolor{orange!25}.8980 & \cellcolor{orange!25}.0539 & \cellcolor{orange!25}23.67 & \cellcolor{orange!25}.7299 & \cellcolor{orange!25}.1082 \\
	        \specialrule{0.6pt}{1pt}{1pt}
			\bottomrule
      \end{tabular}
      }%
   \caption{Quantitative results for the real scene defocus blur. Each color shading indicates the \colorbox{orange!25}{best} and \colorbox{yellow!25}{second-best} result, respectively.}
   \label{tab:quantitative_results_real_defocus}
   \end{center}
\end{table*}

\clearpage


\clearpage


\begin{table*}[t]
   \Large
   \begin{center}
      \resizebox{2.1\columnwidth}{!}{
		\centering
		\setlength{\tabcolsep}{1pt}
         \begin{tabular}{c||ccc|ccc|ccc|ccc|ccc|ccc}
	        \specialrule{0.6pt}{1pt}{1pt}
			\multirow{2}{*}{Camera Motion}& \multicolumn{3}{c}{Factory} & \multicolumn{3}{c}{Cozyroom} & \multicolumn{3}{c}{Pool} & \multicolumn{3}{c}{Tanabata} & \multicolumn{3}{c}{Trolley} & \multicolumn{3}{c}{Average} \\
			 & PSNR($\uparrow$) & SSIM($\uparrow$) & LPIPS($\downarrow$) & PSNR($\uparrow$) & SSIM($\uparrow$) & LPIPS($\downarrow$) & PSNR($\uparrow$) & SSIM($\uparrow$) & LPIPS($\downarrow$) & PSNR($\uparrow$) & SSIM($\uparrow$) & LPIPS($\downarrow$) & PSNR($\uparrow$) & SSIM($\uparrow$) & LPIPS($\downarrow$) & PSNR($\uparrow$) & SSIM($\uparrow$) & LPIPS($\downarrow$) \\
			\midrule
			Naive NeRF~\cite{mildenhall2020nerf} & 19.32 & .4563 & .5304 & 25.66 & .7941 & .2288 & 30.45 & .8354 & .1932 & 22.22 & .6807 & .3653 & 21.25 & .6370 & .3633 & 23.78 & .6807 & .3362 \\
			DP-NeRF(RBK) & \cellcolor{yellow!25}25.82 & \cellcolor{orange!25}.7853 & \cellcolor{yellow!25}.2493 & \cellcolor{orange!25}32.75 & \cellcolor{orange!25}.9331 & \cellcolor{orange!25}.0351 & \cellcolor{orange!25}31.98 & \cellcolor{yellow!25}.8756 & \cellcolor{yellow!25}.0925 & \cellcolor{orange!25}27.63 & \cellcolor{yellow!25}.8745 & \cellcolor{yellow!25}.1045 & \cellcolor{yellow!25}27.93 & \cellcolor{yellow!25}.8723 & \cellcolor{yellow!25}.1130 & \cellcolor{yellow!25}29.22 & \cellcolor{orange!25}.8682 & \cellcolor{yellow!25}.1189 \\
			DP-NeRF(RBK+AWP) &  \cellcolor{orange!25}25.91 & \cellcolor{yellow!25}.7787 & \cellcolor{orange!25}.2494 & \cellcolor{yellow!25}32.65 & \cellcolor{yellow!25}.9317 & \cellcolor{yellow!25}.0355 & \cellcolor{yellow!25}31.96 & \cellcolor{orange!25}.8768 & \cellcolor{orange!25}.0908 & \cellcolor{yellow!25}27.61 & \cellcolor{orange!25}.8748 & \cellcolor{orange!25}.1033 & \cellcolor{orange!25}28.03 & \cellcolor{orange!25}.8752 & \cellcolor{orange!25}.1129 & \cellcolor{orange!25}29.23 & \cellcolor{yellow!25}.8674 & \cellcolor{orange!25}.1184 \\
			\midrule\midrule
			\multirow{2}{*}{Defocus}& \multicolumn{3}{c}{Factory} & \multicolumn{3}{c}{Cozyroom} & \multicolumn{3}{c}{Pool} & \multicolumn{3}{c}{Tanabata} & \multicolumn{3}{c}{Trolley} & \multicolumn{3}{c}{Average} \\
			 & PSNR($\uparrow$) & SSIM($\uparrow$) & LPIPS($\downarrow$) & PSNR($\uparrow$) & SSIM($\uparrow$) & LPIPS($\downarrow$) & PSNR($\uparrow$) & SSIM($\uparrow$) & LPIPS($\downarrow$) & PSNR($\uparrow$) & SSIM($\uparrow$) & LPIPS($\downarrow$) & PSNR($\uparrow$) & SSIM($\uparrow$) & LPIPS($\downarrow$) & PSNR($\uparrow$) & SSIM($\uparrow$) & LPIPS($\downarrow$) \\
			\midrule
			Naive NeRF~\cite{mildenhall2020nerf} & 25.36 & .7847 & .2351 & 30.03 & .8926 & .0885 & 27.77 & .7266 & .3340 & 23.80 & .7811 & .2142 & 22.67 & .7103 & .2799 & 25.93 & .7791 & .2303 \\
			DP-NeRF(RBK) & \cellcolor{yellow!25}28.56 & \cellcolor{yellow!25}.8672 & \cellcolor{yellow!25}.1052 & \cellcolor{yellow!25}32.00 & \cellcolor{yellow!25}.9207 & \cellcolor{yellow!25}.0410 & \cellcolor{yellow!25}31.18 & \cellcolor{yellow!25}.8482 & \cellcolor{yellow!25}.1577 & \cellcolor{yellow!25}26.51 & \cellcolor{yellow!25}.8586 & \cellcolor{yellow!25}.0802 & \cellcolor{yellow!25}26.00 & \cellcolor{yellow!25}.8277 & \cellcolor{yellow!25}.1200 & \cellcolor{yellow!25}28.85 & \cellcolor{yellow!25}.8645 & \cellcolor{yellow!25}.1008 \\
			DP-NeRF(RBK+AWP) & \cellcolor{orange!25}29.26 & \cellcolor{orange!25}.8793 & \cellcolor{orange!25}.1035 & \cellcolor{orange!25}32.11 & \cellcolor{orange!25}.9215 & \cellcolor{orange!25}.0386 & \cellcolor{orange!25}31.44 & \cellcolor{orange!25}.8529 & \cellcolor{orange!25}.1563 & \cellcolor{orange!25}27.05 & \cellcolor{orange!25}.8635 & \cellcolor{orange!25}.0779 & \cellcolor{orange!25}26.79 & \cellcolor{orange!25}.8395 & \cellcolor{orange!25}.1170 & \cellcolor{orange!25}29.33 & \cellcolor{orange!25}.8713 & \cellcolor{orange!25}.0987 \\
			\specialrule{0.6pt}{1pt}{1pt}
			\bottomrule
      \end{tabular}
      }%
   \vspace{-0.1cm}
   \caption{Ablation study for the synthetic scene. Each color shading indicates the \colorbox{orange!25}{best} and \colorbox{yellow!25}{second-best} result, respectively.}
   \label{tab:ablation_synthetic}
   \end{center}
   \vspace{-0.3cm}
\end{table*}
\begin{table*}[t]
   \Large
   \begin{center}
      \resizebox{2.1\columnwidth}{!}{
		\centering
		\setlength{\tabcolsep}{1pt}
         \begin{tabular}{c||ccc|ccc|ccc|ccc|ccc|ccc}
	        \specialrule{0.6pt}{1pt}{1pt}
			\multirow{2}{*}{Camera Motion}& \multicolumn{3}{c}{Ball} & \multicolumn{3}{c}{Basket} & \multicolumn{3}{c}{Buick} & \multicolumn{3}{c}{Coffee} & \multicolumn{3}{c}{Decoration} & & \\
			 & PSNR($\uparrow$) & SSIM($\uparrow$) & LPIPS($\downarrow$) & PSNR($\uparrow$) & SSIM($\uparrow$) & LPIPS($\downarrow$) & PSNR($\uparrow$) & SSIM($\uparrow$) & LPIPS($\downarrow$) & PSNR($\uparrow$) & SSIM($\uparrow$) & LPIPS($\downarrow$) & PSNR($\uparrow$) & SSIM($\uparrow$) & LPIPS($\downarrow$) &  & &\\
			\midrule
			Naive NeRF~\cite{mildenhall2020nerf} & 24.08 & .6237 & .3992 & 23.72 & .7086 & .3223 & 21.59 & .6325 & .3502 & 26.48 & .8064 & .2896 & 22.39 & .6609 & .3633 & & & \\
			DP-NeRF(RBK) & \cellcolor{yellow!25}27.15 & \cellcolor{yellow!25}.7641 & \cellcolor{yellow!25}.2112 & \cellcolor{yellow!25}27.35 & \cellcolor{yellow!25}.8367 & \cellcolor{yellow!25}.1347 & \cellcolor{yellow!25}24.93 & \cellcolor{yellow!25}.7791 & \cellcolor{yellow!25}.1545 & \cellcolor{yellow!25}30.72 & \cellcolor{yellow!25}.8949 & \cellcolor{yellow!25}.1070 & \cellcolor{yellow!25}24.15 & \cellcolor{yellow!25}.7730 & \cellcolor{yellow!25}.1700 & & & \\
			DP-NeRF(RBK+AWP):& \cellcolor{orange!25}27.20 & \cellcolor{orange!25}.7652 & \cellcolor{orange!25}.2088 & \cellcolor{orange!25}27.74 & \cellcolor{orange!25}.8455 & \cellcolor{orange!25}.1294 & \cellcolor{orange!25}25.70 & \cellcolor{orange!25}.7922 & \cellcolor{orange!25}.1405 & \cellcolor{orange!25}31.19 & \cellcolor{orange!25}.9049 & \cellcolor{orange!25}.1002 & \cellcolor{orange!25}24.31 & \cellcolor{orange!25}.7811 & \cellcolor{orange!25}.1639 & & & \\
			\midrule\midrule
			\multirow{2}{*}{Camera motion}& \multicolumn{3}{c}{Girl} & \multicolumn{3}{c}{Heron} & \multicolumn{3}{c}{Parterre} & \multicolumn{3}{c}{Puppet} & \multicolumn{3}{c}{Stair} & \multicolumn{3}{c}{Average} \\
			 & PSNR($\uparrow$) & SSIM($\uparrow$) & LPIPS($\downarrow$) & PSNR($\uparrow$) & SSIM($\uparrow$) & LPIPS($\downarrow$) & PSNR($\uparrow$) & SSIM($\uparrow$) & LPIPS($\downarrow$) & PSNR($\uparrow$) & SSIM($\uparrow$) & LPIPS($\downarrow$) & PSNR($\uparrow$) & SSIM($\uparrow$) & LPIPS($\downarrow$) & PSNR($\uparrow$) & SSIM($\uparrow$) & LPIPS($\downarrow$) \\
			\midrule
			Naive NeRF~\cite{mildenhall2020nerf} & 20.07 & .7075 & .3196 & 20.50 & .5217 & .4129 & 23.14 & .6201 & .4046 & 22.09 & .6093 & .3389 & 22.87 & .4561 & .4868 & 22.69 & .6347 & .3687 \\
			DP-NeRF(RBK) & \cellcolor{yellow!25}22.19 & \cellcolor{yellow!25}.7934 & \cellcolor{yellow!25}.1575 & \cellcolor{yellow!25}22.55 & \cellcolor{yellow!25}.6831 & \cellcolor{yellow!25}.1970 & \cellcolor{yellow!25}25.81 & .\cellcolor{yellow!25}7635 & \cellcolor{yellow!25}.1931 & \cellcolor{yellow!25}25.19 & \cellcolor{yellow!25}.7497 & \cellcolor{orange!25}.1493 & \cellcolor{orange!25}25.68 & \cellcolor{orange!25}.6446 & \cellcolor{yellow!25}.1799 & \cellcolor{yellow!25}25.57 & \cellcolor{yellow!25}.7682 & \cellcolor{yellow!25}.1654 \\
			DP-NeRF(RBK+AWP) & \cellcolor{orange!25}23.33 & \cellcolor{orange!25}.8139 & \cellcolor{orange!25}.1498 & \cellcolor{orange!25}22.88 & \cellcolor{orange!25}.6930 & \cellcolor{orange!25}.1914 & \cellcolor{orange!25}25.86 & \cellcolor{orange!25}.7665 & \cellcolor{orange!25}.1900 & \cellcolor{orange!25}25.25 & \cellcolor{orange!25}.7536 & \cellcolor{yellow!25}.1505 & \cellcolor{yellow!25}25.59 & \cellcolor{yellow!25}.6349 & \cellcolor{orange!25}.1772 & \cellcolor{orange!25}25.91 & \cellcolor{orange!25}.7751 & \cellcolor{orange!25}.1602 \\
	        \specialrule{0.6pt}{1pt}{1pt}
			\midrule
      \end{tabular}
      }%
   \vspace{-0.1cm}
   \caption{Ablation study for the real scene camera motion blur. Each color shading indicates the \colorbox{orange!25}{best} and \colorbox{yellow!25}{second-best} result, respectively.}
   \label{tab:ablation_real_camera_motion}
   \end{center}
   \vspace{-0.3cm}
\end{table*}
\begin{table*}[!t]
   \Large
   \begin{center}
      \resizebox{2.1\columnwidth}{!}{
		\centering
		\setlength{\tabcolsep}{1pt}
         \begin{tabular}{c||ccc|ccc|ccc|ccc|ccc|ccc}
	        \specialrule{0.6pt}{1pt}{1pt}
			\multirow{2}{*}{Defocus}& \multicolumn{3}{c}{Cake} & \multicolumn{3}{c}{Caps} & \multicolumn{3}{c}{Cisco} & \multicolumn{3}{c}{Cupcake} & \multicolumn{3}{c}{Coral} & \multicolumn{3}{c}{}\\
			 & PSNR($\uparrow$) & SSIM($\uparrow$) & LPIPS($\downarrow$) & PSNR($\uparrow$) & SSIM($\uparrow$) & LPIPS($\downarrow$) & PSNR($\uparrow$) & SSIM($\uparrow$) & LPIPS($\downarrow$) & PSNR($\uparrow$) & SSIM($\uparrow$) & LPIPS($\downarrow$) & PSNR($\uparrow$) & SSIM($\uparrow$) & LPIPS($\downarrow$) & & & \\
			\midrule
			Naive NeRF~\cite{mildenhall2020nerf} & 24.42 & .7210 & .2250 & 22.73 & .6312 & .2801 & 20.72 & .7217 & .1256 & 21.88 & .6809 & .2155 & 19.81 & .5658 & .2689 & & & \\
			DP-NeRF(RBK) & \cellcolor{yellow!25}25.80 & \cellcolor{yellow!25}.7704 & \cellcolor{orange!25}.1251 & \cellcolor{yellow!25}23.72 & \cellcolor{yellow!25}.7003 & \cellcolor{yellow!25}.1486 & \cellcolor{yellow!25}20.68 & \cellcolor{yellow!25}.7232 & \cellcolor{yellow!25}.0889 & \cellcolor{yellow!25}22.51 & \cellcolor{yellow!25}.7331 & \cellcolor{yellow!25}.1003 & \cellcolor{yellow!25}20.02 & \cellcolor{yellow!25}.6052 & \cellcolor{yellow!25}.1183  & & & \\
			DP-NeRF(RBK+AWP) & \cellcolor{orange!25}26.16 & \cellcolor{orange!25}.7781 & \cellcolor{yellow!25}.1267 & \cellcolor{orange!25}23.95 & \cellcolor{orange!25}.7122 & \cellcolor{orange!25}.1430 & \cellcolor{orange!25}20.73 & \cellcolor{orange!25}.7260 & \cellcolor{orange!25}.0840 & \cellcolor{orange!25}22.80 & \cellcolor{orange!25}.7409 & \cellcolor{orange!25}.0960 & \cellcolor{orange!25}20.11 & \cellcolor{orange!25}.6107 & \cellcolor{orange!25}.1178 & & & \\
			\midrule\midrule
			\multirow{2}{*}{Defocus}& \multicolumn{3}{c}{Cups} & \multicolumn{3}{c}{Daisy} & \multicolumn{3}{c}{Sausage} & \multicolumn{3}{c}{Seal} & \multicolumn{3}{c}{Tools} & \multicolumn{3}{c}{Average} \\
			 & PSNR($\uparrow$) & SSIM($\uparrow$) & LPIPS($\downarrow$) & PSNR($\uparrow$) & SSIM($\uparrow$) & LPIPS($\downarrow$) & PSNR($\uparrow$) & SSIM($\uparrow$) & LPIPS($\downarrow$) & PSNR($\uparrow$) & SSIM($\uparrow$) & LPIPS($\downarrow$) & PSNR($\uparrow$) & SSIM($\uparrow$) & LPIPS($\downarrow$) & PSNR($\uparrow$) & SSIM($\uparrow$) & LPIPS($\downarrow$) \\
			\midrule
			Naive NeRF~\cite{mildenhall2020nerf} & 25.02 & .7581 & .2315 & 22.74 & .6203 & .2621 & 17.79 & .4830 & .2789 & 22.79 & .6267 & .2680 & 26.08 & .8523 & .1547 & 22.40 & .6661 & .2310 \\
			DP-NeRF(RBK) & \cellcolor{yellow!25}26.59 & \cellcolor{yellow!25}.8086 & \cellcolor{yellow!25}.1077 & \cellcolor{yellow!25}23.77 & \cellcolor{yellow!25}.6968 & \cellcolor{orange!25}.1059 & \cellcolor{orange!25}18.40 & \cellcolor{orange!25}.5448 & \cellcolor{orange!25}.1452 & \cellcolor{orange!25}26.04 & \cellcolor{yellow!25}.7767 & \cellcolor{orange!25}.0996 & \cellcolor{yellow!25}27.87 & \cellcolor{yellow!25}.8947 & \cellcolor{yellow!25}.0540 & \cellcolor{yellow!25}23.54 & \cellcolor{yellow!25}.7254 & \cellcolor{yellow!25}.1093 \\
			DP-NeRF(RBK+AWP) & \cellcolor{orange!25}26.75 & \cellcolor{orange!25}.8136 & \cellcolor{orange!25}.1035 & \cellcolor{orange!25}23.79 & \cellcolor{orange!25}.6971 & \cellcolor{yellow!25}.1075 &  \cellcolor{yellow!25}18.35 &  \cellcolor{yellow!25}.5443 &  \cellcolor{yellow!25}.1473 &  \cellcolor{yellow!25}25.95 &  \cellcolor{orange!25}.7779 &  \cellcolor{yellow!25}.1026 & \cellcolor{orange!25}28.07 & \cellcolor{orange!25}.8980 & \cellcolor{orange!25}.0539 & \cellcolor{orange!25}23.67 & \cellcolor{orange!25}.7299 & \cellcolor{orange!25}.1082 \\
	        \specialrule{0.6pt}{1pt}{1pt}
			\bottomrule
      \end{tabular}
      }%
   \vspace{-0.1cm}
   \caption{Ablation study for the real scene defocus blur. Each color shading indicates the \colorbox{orange!25}{best} and \colorbox{yellow!25}{second-best} result, respectively.}
   \label{tab:ablation_real_defocus}
   \end{center}
   \vspace{-0.5cm}
\end{table*}

\clearpage

\begin{figure}[t!]
\footnotesize
      \begin{center}
         \includegraphics[scale=1.0]{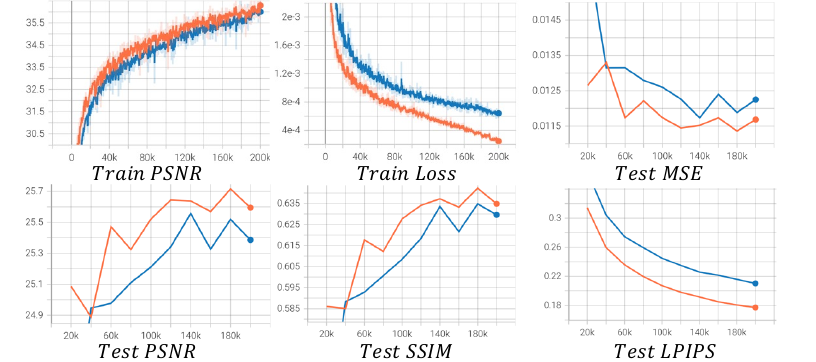} 
      \caption{Training of \textcolor{orange}{DP-NeRF} and \textcolor{blue}{Deblur-NeRF} on \textit{Stair} scene.}
      \label{fig:opt_chart}
      \vspace{-0.5cm}
      \end{center}
\end{figure}

\begin{table}[!t]
   \begin{center}
   \footnotesize
      \resizebox{\columnwidth}{!}{
		\centering
         \begin{tabular}{c||ccc|}
	        \toprule
			Model & PSNR($\uparrow$) & SSIM($\uparrow$) & LPIPS($\downarrow$) \\
			\midrule
			DP-NeRF(RBK) & 24.93 & .7791 & .1545 \\
			DP-NeRF(RBK+AWP) & \cellcolor{yellow!25}25.47 & \cellcolor{yellow!25}.7883 & \cellcolor{yellow!25}.1466 \\
			DP-NeRF(RBK+AWP+Coarse-to-Fine) & \cellcolor{orange!25}25.70 & \cellcolor{orange!25}.7922 & \cellcolor{orange!25}.1405 \\
			\bottomrule
		\end{tabular}
      	}
   \vspace{-0.2cm}
   \caption{Ablation of coarse-to-fine optimization on \textit{Buick} scene.}
   \label{tab:abl_c2f}
   \end{center}
   \vspace{-0.2cm}
\end{table}

\subsection{Ablation of Coarse to Fine Optimization}
\label{sub:c2f_opt}
We present the optimization graph for Deblur-NeRF~\cite{ma2022deblurnerf} and DP-NeRF in \figurename~\ref{fig:opt_chart}.
Though it seems to be harder to optimize the DP-NeRF due to higher degree of freedom, DP-NeRF converges more faster than Deblur-NeRF thanks to the shared rigid motions in a single view, which works as regularization to optimize the RBK. 
In addition, we attach ablation results of the coarse-to-fine optimization in \tablename~\ref{tab:abl_c2f}. 
It reveals that the proposed optimization scheme helps the model to take the full advantage of AWP and improves the visual quality. 
The time required to train $\textit{\textbf{200000}}$ iterations are around \textit{\textbf{9.5 hours}} for Deblur-NeRF and \textit{\textbf{19 hours}} for DP-NeRF on \textit{two NVIDIA RTX 3090 GPUs}. 
 Note that, we present representative scenes for \figurename~\ref{fig:opt_chart} and \tablename~\ref{tab:abl_c2f} due to time limitation to experiment on all scenes.

\subsection{Ablation of the Number of Rigid Motions}
\label{sub:quantative_abl}
\begin{figure*}[t]
      \centering
         \includegraphics[scale=1.0]{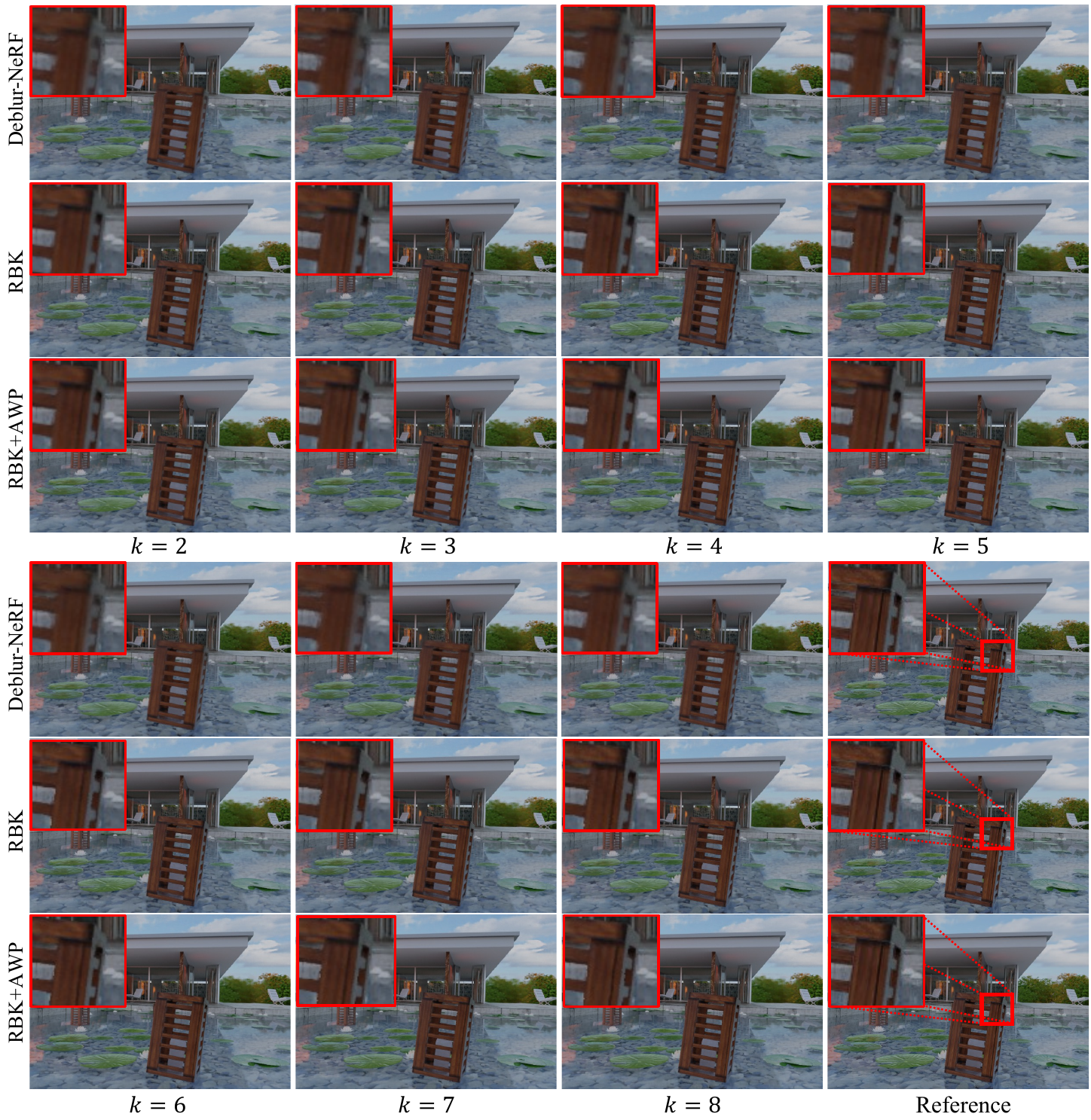} 
      \vspace{-0.3cm}
      \caption{Qualitative results of ablation study depending on the number of rigid motion on camera motion blur scene. $k$ denotes the number of rigid motion. Each row denotes Deblur-NeRF, DP-NeRF(RBK) and DP-NeRF(RBK+AWP), respectively. \textcolor{red}{Red} colored box in corner of images are enlarged part of same colored box region in reference images.}
      \vspace{-0.2cm}
      \label{fig:motion_abls_qualitative_camera_motion}
   \end{figure*}

\begin{figure*}[t]
      \centering
         \includegraphics[scale=1.0]{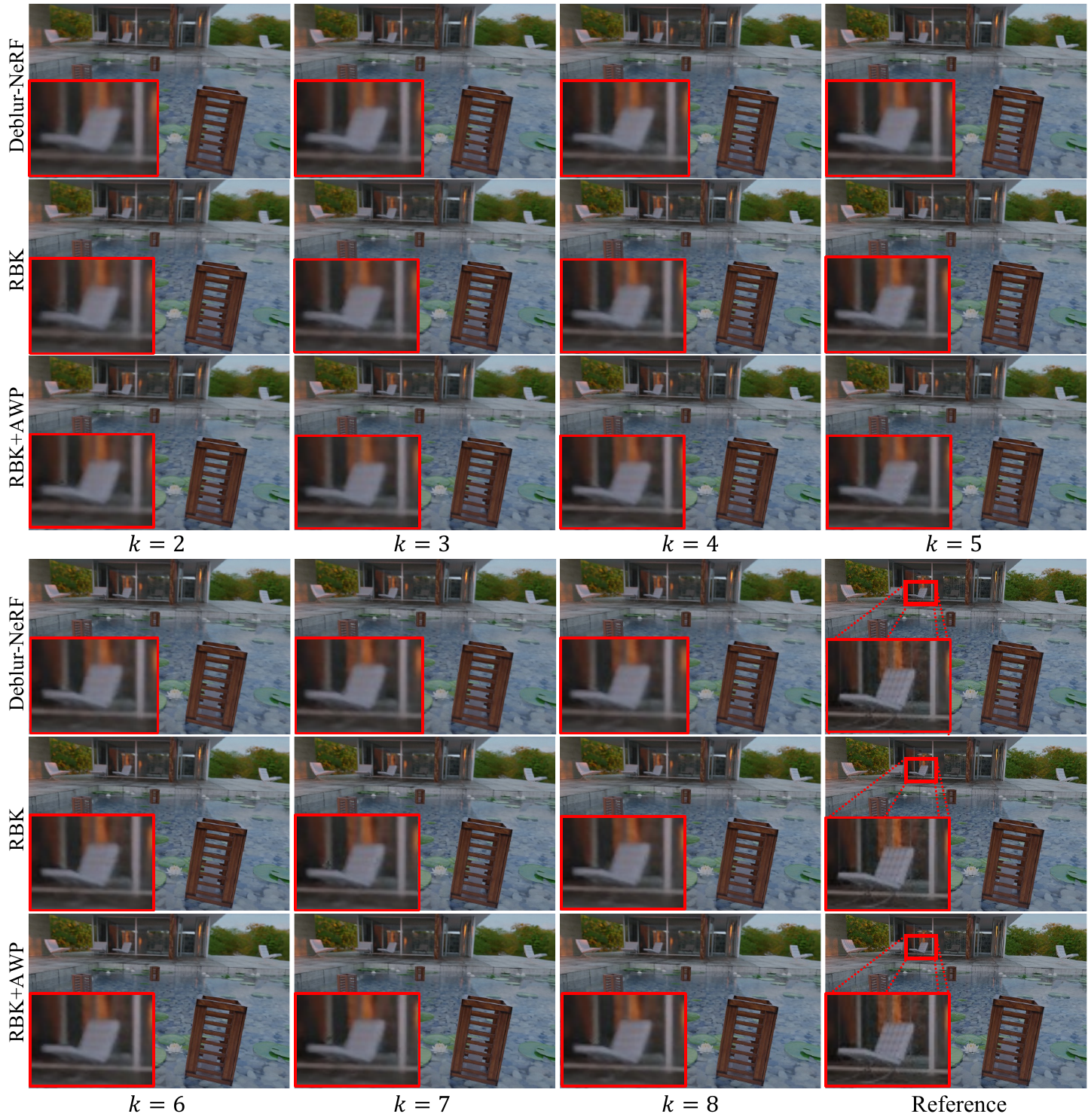} 
      \vspace{-0.3cm}
      \caption{Qualitative results of ablation study depending on the number of rigid motion on defocus blur scene. $k$ denotes the number of rigid motion. Each row denotes Deblur-NeRF, DP-NeRF(RBK) and DP-NeRF(RBK+AWP), respectively. \textcolor{red}{Red} colored box in corner of images are enlarged part of same colored box region in reference images.}
      \vspace{-0.3cm}
      \label{fig:motion_abls_qualitative_defocus}
   \end{figure*}

We present the quantitative and qualitative results for the ablation analysis of the number of rigid motions in \tablename~\ref{tab:abls_quantitative_main_paper} and Figures~\ref{fig:motion_abls_qualitative_camera_motion} and \ref{fig:motion_abls_qualitative_defocus}.
As we mentioned, LPIPS represents perceptual quality better than PSNR and SSIM do.
Based on this metric, the perceptual quality of DP-NERF gradually improves as the number of rigid motions increases.
In addition, DP-NeRF exhibits higher perceptual quality than Deblur-NeRF~\cite{ma2022deblurnerf} in terms of LPIPS for the same number of composition rays.
Furthermore, RBK+AWP produces a better performance than does the RBK alone.

\subsection{Visualization of Rigid Transformed Rays}
\label{sub:RT_rays_vis}
\begin{figure*}[t]
      \centering
         \includegraphics[scale=1.15]{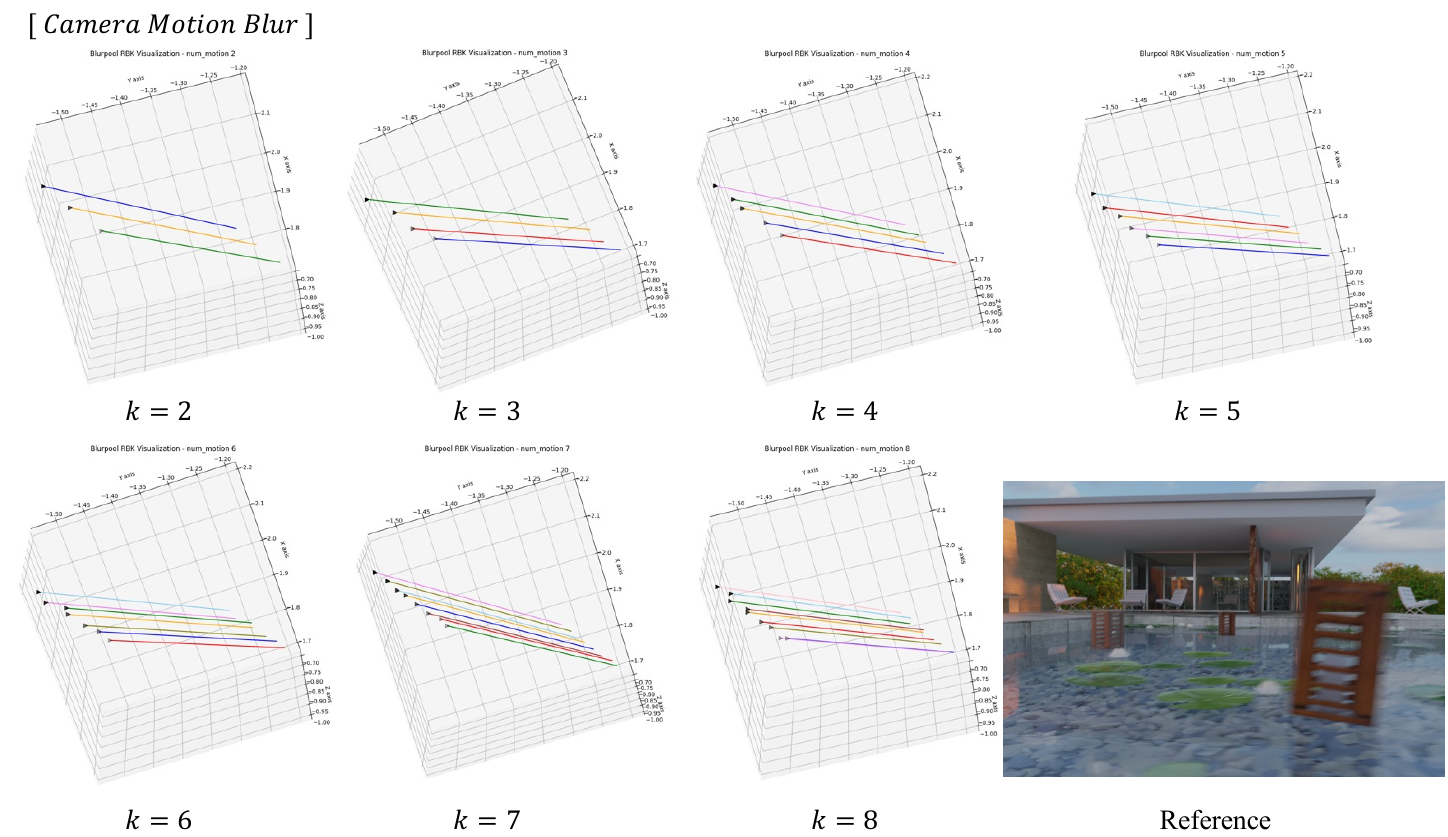} 
      \caption{Visualization of warped poses using rigid motions predicted by the RBK according to the change of the number of rigid motion $k$ on camera motion blur.}
      \vspace{-0.3cm}
      \label{fig:vis_RT_rays_camera_motion}
   \end{figure*}

\begin{figure*}[t]
      \centering
         \includegraphics[scale=1.15]{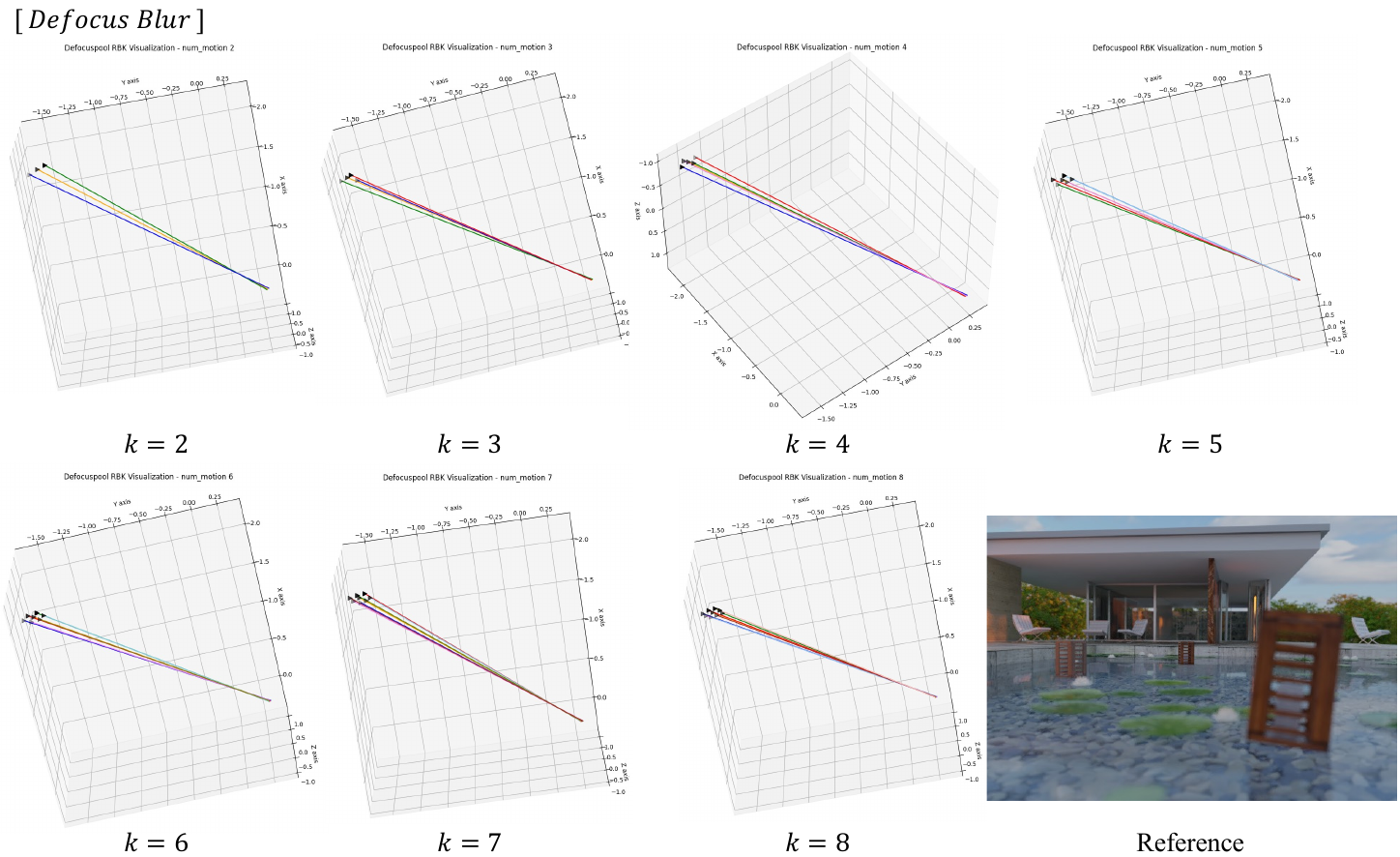} 
      \caption{Visualization of warped poses using rigid motions predicted by the RBK according to the change of the number of rigid motion $k$ on defocus blur.}
      \vspace{-0.3cm}
      \label{fig:vis_RT_rays_defocus}
   \end{figure*}

We present additional visualization of rigid transformed rays predicted by the RBK for a specific view.
\figurename~\ref{fig:vis_RT_rays_camera_motion} and \ref{fig:vis_RT_rays_defocus} show the transformed rays according to the change of the number of rigid motions $k$ for camera motion and defocus blur, respectively. The rays colored as \textcolor{orange!75}{orange} are original center rays in all visualization, which are forwarded to RBK to generate the other rigidly warped rays in both \figurename~\ref{fig:vis_RT_rays_camera_motion} and \ref{fig:vis_RT_rays_defocus}. They demonstrate that RBK successfully model the camera shaking and virtual aperture for camera motion and defocus blur, respectively.

\subsection{Additional Error Map Visualization}
\label{sub:qualitative_error_map}
\begin{figure*}[t]
      \centering
         \includegraphics[scale=1.0]{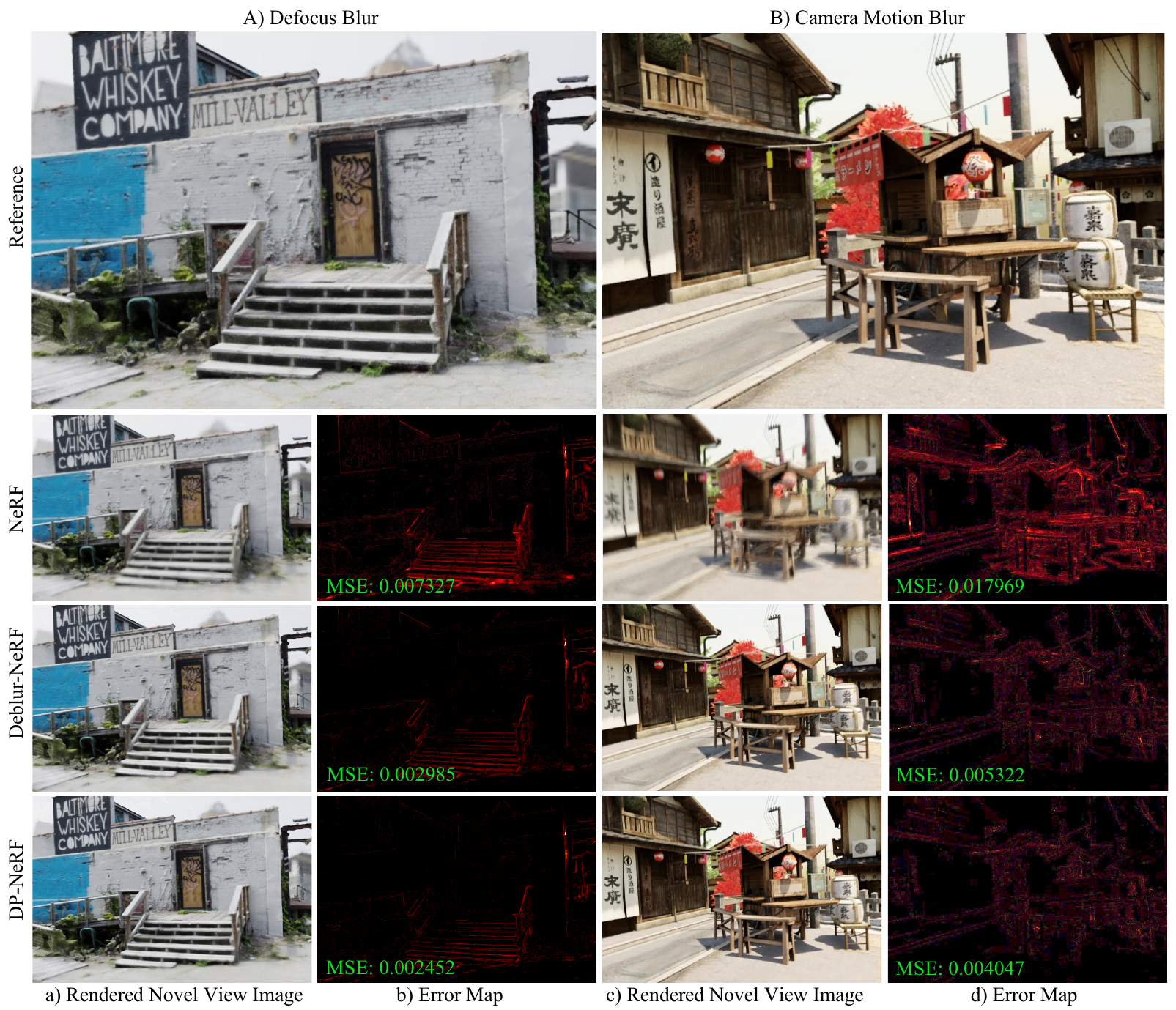} 
      \caption{Full error map visualization on defocus \textbf{Factory} and camera motion \textbf{Trolley} scene. In addition, we denote quantitative quality between reference and rendered images as average MSE(Mean Squared Error), denoted as \textcolor{green}{green text} on left bottom of each images.}
      \vspace{-0.3cm}
      \label{fig:full_error_map}
   \end{figure*}

As mentioned in the main paper, we present a full error map visualization without the emphasis box in \figurename~\ref{fig:full_error_map}.
Note that, brighter colors indicate greater error.
We control the brightness and contrast in all output identically, to reveal the error between the models more prominently.
Our model produces less error than the baselines for the images regardless of the blur type.
\vspace{-0.2cm}

\section{Supplementary Video}
\label{sec:supple_video}
We attach videos outlining novel view synthesis.
These videos are generated with spiral-path camera poses, which are widely used for visual comparison in NeRF-based research.
Please watch theses supplementary videos or visit out project page to observe the qualitative effectiveness of DP-NERF.

\clearpage

\begin{table*}[!t]
   \Huge
   \begin{center}
      \resizebox{2.1\columnwidth}{!}{
		\centering
		\setlength{\tabcolsep}{1pt}
         \begin{tabular}{c||c|c|c|c|c|c|c|c|c||c|c|c|c|c|c|c|c|c}
	        \specialrule{0.6pt}{1pt}{1pt}
			& \multicolumn{9}{c}{Camera Motion - Pool} &  \multicolumn{9}{c}{Defocus - Pool} \\
			\midrule
			\multirow{2}{*}{~\# of RM~}& \multicolumn{3}{c}{DeblurNeRF} & \multicolumn{3}{c}{DP-NeRF(RBK)} & \multicolumn{3}{c}{DP-NeRF(RBK+AWP)} & \multicolumn{3}{c}{DeblurNeRF} & \multicolumn{3}{c}{DP-NeRF(RBK)} & \multicolumn{3}{c}{DP-NeRF(RBK+AWP)} \\
			 & PSNR($\uparrow$) & SSIM($\uparrow$) & LPIPS($\downarrow$) & PSNR($\uparrow$) & SSIM($\uparrow$) & LPIPS($\downarrow$) & PSNR($\uparrow$) & SSIM($\uparrow$) & LPIPS($\downarrow$) & PSNR($\uparrow$) & SSIM($\uparrow$) & LPIPS($\downarrow$) & PSNR($\uparrow$) & SSIM($\uparrow$) & LPIPS($\downarrow$) & PSNR($\uparrow$) & SSIM($\uparrow$) & LPIPS($\downarrow$)\\
			\midrule
			2 & 31.48 & .8658 & .1348 & \cellcolor{yellow!25}31.97 & \cellcolor{yellow!25}.8741 & .1102 & \cellcolor{orange!25}32.30 & \cellcolor{orange!25}.8822 & .1031 & 29.89 & .8087 & .2280 & 30.24 & .8246 & .2023 & 30.52 & .8301 & .1971 \\
			3 & 31.52 & .8670 & .1289 & 31.61 & .8682 & .1032 & 31.80 & .8709 & .0966 & 30.07 & .8117 & .2094 & 29.22 & .7923 & .1891 & 29.39 & .8006 & .1871 \\
			4 & 31.49 & .8672 & .1257 & \cellcolor{orange!25}31.98 & \cellcolor{orange!25}.8756 & .0925 & \cellcolor{yellow!25}31.96 & \cellcolor{yellow!25}.8768 & .0908 & 30.43 & .8230 & .1931 & 31.18 & .8482 & .1577 & 31.44 & .8529 & .1563 \\
			5 & \cellcolor{orange!25}31.59 & \cellcolor{orange!25}.8710 & .1169 & 31.40 & .8671 & .0912 & 31.88 & .8763 & .0849 & 30.52 & .8244 & .1832 & 30.99 & .8421 & .1485 & 31.07 & .8473 & .1484\\
			6 & \cellcolor{orange!25}31.59 & \cellcolor{yellow!25}.8688 & .1149 & 31.43 & .8662 & .0868 & 31.74 & .8733 & .0807 & 30.70 & .8310 & .1712 & \cellcolor{orange!25}31.42 & \cellcolor{orange!25}.8549 & .1368 & \cellcolor{yellow!25}31.73 & \cellcolor{orange!25}.8621& .1362 \\
			7 & 31.42 & .8628 & \cellcolor{yellow!25}.1121 & 31.65 & .8699 & \cellcolor{yellow!25}.0851 & 31.67 & .8714 & \cellcolor{yellow!25}.0806 & \cellcolor{yellow!25}30.73 & \cellcolor{yellow!25}.8317 & \cellcolor{yellow!25}.1728 & 31.15 & .8484 & \cellcolor{yellow!25}.1328 & 31.62 & .8589 & \cellcolor{yellow!25}.1302 \\
			8 & \cellcolor{yellow!25}31.56 & .8685 & \cellcolor{orange!25}.1097 & 31.49 & .8683 & \cellcolor{orange!25}.0840 & 31.49 & .8668 & \cellcolor{orange!25}.0796 & \cellcolor{orange!25}30.92 & \cellcolor{orange!25}.8352 & \cellcolor{orange!25}.1625 & \cellcolor{yellow!25}31.37 & \cellcolor{yellow!25}.8532 & \cellcolor{orange!25}.1267 & \cellcolor{orange!25}31.77 & \cellcolor{yellow!25}.8609 & \cellcolor{orange!25}.1234 \\
	        \specialrule{0.6pt}{1pt}{1pt}
			\bottomrule
      \end{tabular}
      }%
   \vspace{-0.3cm}
   \caption{Quantitative results for ablation study of the number of rigid motions in the main paper, which is denoted as RM in the table. Each color shading indicates \colorbox{orange!25}{best} and \colorbox{yellow!25}{second-best} result on each scene with each model, respectively. Note that, the number of kernel points in Deblur-NeRF is set to (the number of rigid motions + 1) for fair comparison, which means the same number of composition rays to create a blurred color for a pixel.}
   \label{tab:abls_quantitative_main_paper}
   \end{center}
   \vspace{-0.5cm}
\end{table*}

\clearpage
{\small
\bibliographystyle{ieee_fullname}
\bibliography{egbib}
}

\clearpage
\end{document}